\documentclass[11pt,a4paper]{article}

\usepackage[margin=2.5cm]{geometry}
\usepackage{setspace}
\usepackage{fancyhdr}
\usepackage{environ}
\newcommand{\acksection}{\section*{Acknowledgments}}
\NewEnviron{ack}{
  \acksection
  \BODY
}

\usepackage[utf8]{inputenc}
\usepackage[T1]{fontenc}
\usepackage{hyperref}
\usepackage{url}
\newcommand{\orcidlink}[1]{\href{https://orcid.org/#1}{\textcolor{green!50!black}{ORCID}}}
\usepackage{booktabs}
\usepackage{amsfonts}
\usepackage{nicefrac}
\usepackage{microtype}
\usepackage{xcolor}
\usepackage[round]{natbib}
\usepackage{graphicx}
\usepackage{amsmath}
\usepackage{subcaption}

\usepackage{algorithm}
\usepackage{algpseudocode}

\usepackage{tikz}
\usetikzlibrary{shapes.geometric, arrows.meta, positioning, fit, backgrounds, calc}

\newcommand{\vect}[1]{\mathbf{#1}}
\newcommand{\vx}{\vect{x}}
\newcommand{\vepsilon}{\boldsymbol{\epsilon}}

\newcommand{\vm}{\vect{m}}
\newcommand{\vI}{\vect{I}}
\newcommand{\vz}{\vect{z}}
\newcommand{\vmu}{\vect{\mu}}
\newcommand{\vSigma}{\vect{\Sigma}}
\newcommand{\Normal}{\mathcal{N}}
\newcommand{\E}{\mathbb{E}}

\graphicspath{{plots/}}

\title{Fast 3D Diffusion for Scalable Discrete Granular Media Synthesis\thanks{Source code: \url{https://github.com/FammasMaz/fast-gran-gen}}}

\author{
  Muhammad M. Hassan$^{1,3}$\textsuperscript{\orcidlink{0009-0005-5260-8813}}, Régis Cottereau$^1$\textsuperscript{\orcidlink{0000-0003-0739-6805}}, Filippo Gatti$^2$\textsuperscript{\orcidlink{0000-0001-7174-4048}}, Patryk Dec$^3$\textsuperscript{\orcidlink{0000-0002-1703-7091}} \\
  \\
  $^1$Aix Marseille Univ, CNRS, Centrale Med, LMA, Marseille, France, \\
  4, impasse Nikola Tesla, Marseille, France \\
  $^2$Université Paris Saclay, CentraleSupélec, ENS Paris-Saclay, CNRS, \\
  Laboratoire de Mécanique Paris-Saclay UMR 9026, 91190 Gif-sur-Yvette, France \\
  $^3$Innovation and Research Department, SNCF, 93210 La Plaine Saint Denis, France \\
  \\
  Corresponding author: muhammadmoeeze.hassan@sncf.fr
}

\begin{document}

\maketitle

\begin{abstract}
Discrete Element Method (DEM) simulations of granular media are computationally intensive, particularly during initialization phases dominated by large displacements and kinetic energy. This paper presents a novel generative pipeline based on 3D diffusion models that directly synthesizes arbitrarily large granular assemblies in mechanically realistic configurations. The approach employs a two-stage pipeline. First, an unconditional diffusion model generates independent 3D voxel grids representing granular media; second, a 3D inpainting model, adapted from 2D techniques using masked inputs and repainting strategies, seamlessly stitches these grids together. The inpainting model uses the outputs of the unconditional diffusion model to learn from the context of adjacent generations and creates new regions that blend smoothly into the context region. Both models are trained on binarized 3D occupancy grids derived from a database of small-scale DEM simulations, scaling linearly with the number of output voxels. Simulations that spanned over days can now run in hours, practically enabling simulations containing more than 200k ballast particles. The pipeline remains fully compatible with existing DEM workflows as it post-processes the diffusion generated voxel grids into DEM compatible particle meshes. Being mechanically consistent on key granulometry metrics with the original DEM simulations, the pipeline is also compatible with many other applications in the field of granular media, with capability of generating both convex and non-convex particles. Showcased on two examples (railway ballast and lunar regolith), the pipeline reimagines the way initialization of granular media simulations is performed, enabling scales of generation previously unattainable with traditional DEM simulations.
\end{abstract}


\section{Introduction}
\label{sec:introduction}
Simulation of granular media is crucial across various scientific and industrial domains, yet it often presents significant computational challenges due to the discrete nature of the materials involved \citep{osullivanParticulateDiscreteElement2011,cundallDiscreteNumericalModel1979}. This work introduces a novel deep learning approach to accelerate the generation of realistic granular assemblies by leveraging a class of generative models, called diffusion models \citep{hoDenoisingDiffusionProbabilistic2020}.

\subsection{Motivation and Problem Statement}
\label{subsec:motivation_problem_statement}
Granular media are ubiquitous in nature, and their complex simulations range from mechanical simulations of sand \citep{osullivanParticulateDiscreteElement2011}, to ballasted railway tracks in the transport industry \citep{indraratnaCurrentResearchBallasted2017}. The peculiar discrete nature of granular media, make them often significantly more expensive to simulate compared to their continuum counterparts. The traditional routines used in simulation of these granular media rely on Discrete Element Method (DEM) \citep{kleinertModelingLargeScale2013,osullivanParticulateDiscreteElement2011,cundallDiscreteNumericalModel1979}, which despite their high accuracy in modeling the contact behavior in grains, are inherently computationally intensive and difficult to parallelize \citep{fredericduboisLMGC902013}. There is a broad literature on DEM for ballast and railway applications highlighting these computational challenges (Section~\ref{subsec:dem_init} provides a detailed overview). Most notably, amongst these challenges, the time scale and the sheer number of particle contacts needed to be resolved at high velocities stands out. In most DEM simulations, however, the computational challenge extends beyond the physics solver itself. Simulations need an initialization phase where the sample on which the simulation is to be run, needs to be setup in a specific state that conforms to realistic situations. This initialization phase constitutes the key computational bottleneck that this paper addresses, i.e., it typically consumes a significant portion of the total simulation time because particles must travel large distances while the solver adopts very small time steps to accurately model rapidly changing contacts when the grains are moving at high speeds.

\begin{figure}[htb]
	\centering
	\begin{subfigure}[htb]{0.49\textwidth}
		\includegraphics[width=\textwidth]{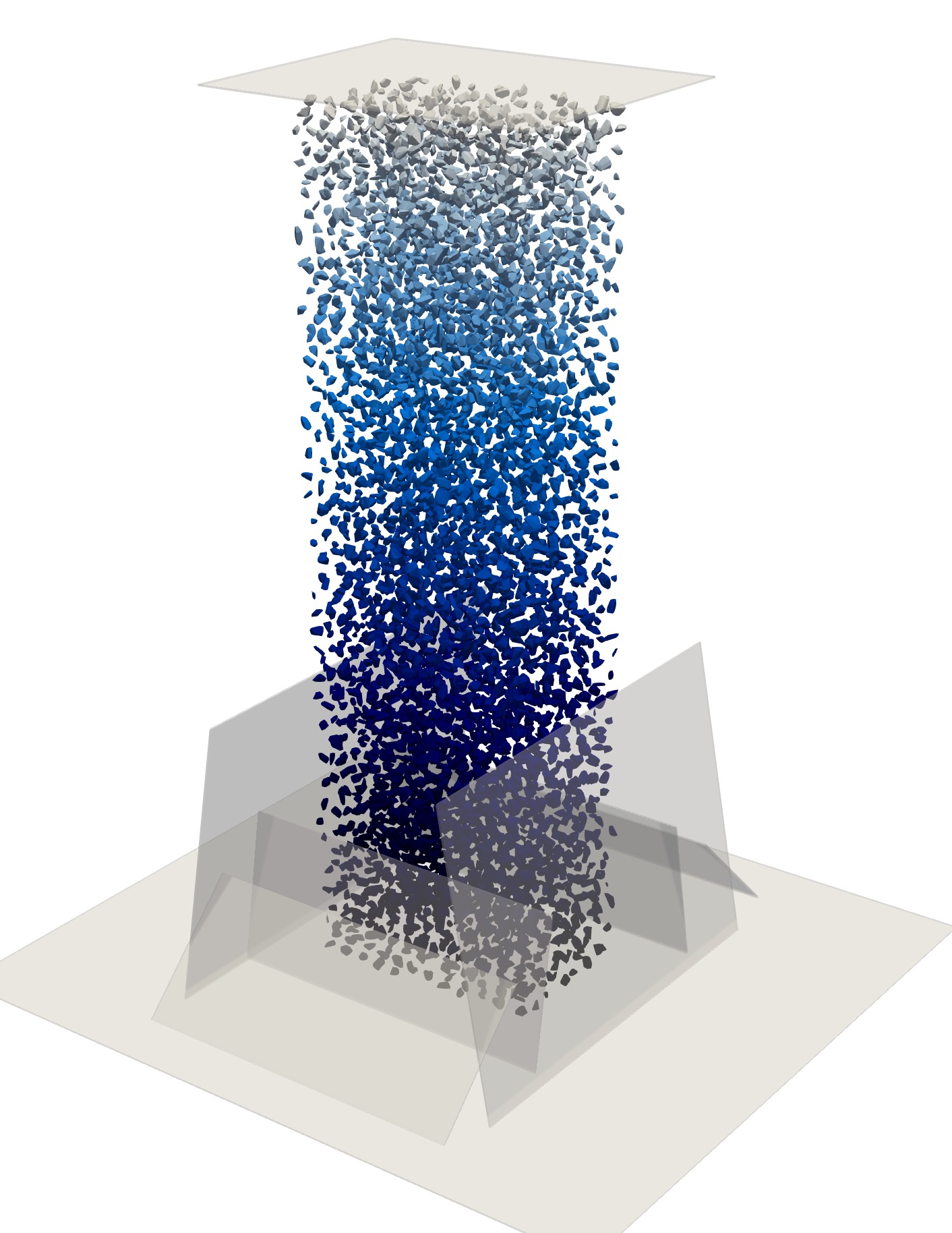}
		\caption{Inital State}
		\label{fig:fall_first.jpg}
	\end{subfigure}
	\begin{subfigure}[htb]{0.49\textwidth}
		\includegraphics[width=\textwidth]{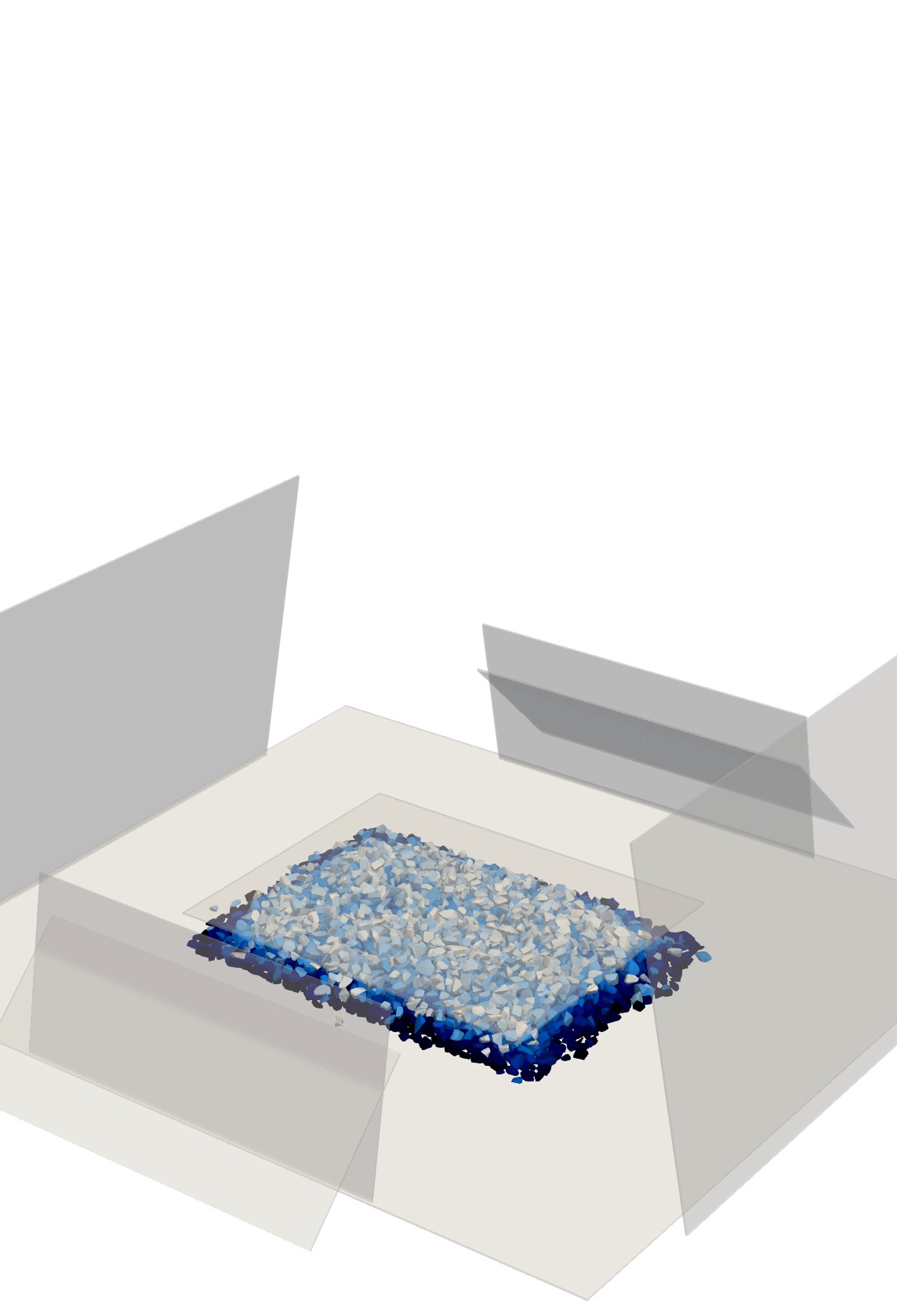}
		\caption{Final State}
		\label{fig:fall_final.jpg}
	\end{subfigure}
	\caption{A typical DEM simulation to create a stack of ballast particles for a railway simulation.}
	\label{fig:fall_first.jpg-fall_final.jpg}
\end{figure}

\begin{figure}[htp]
	\centering
	\includegraphics[width=0.99\textwidth]{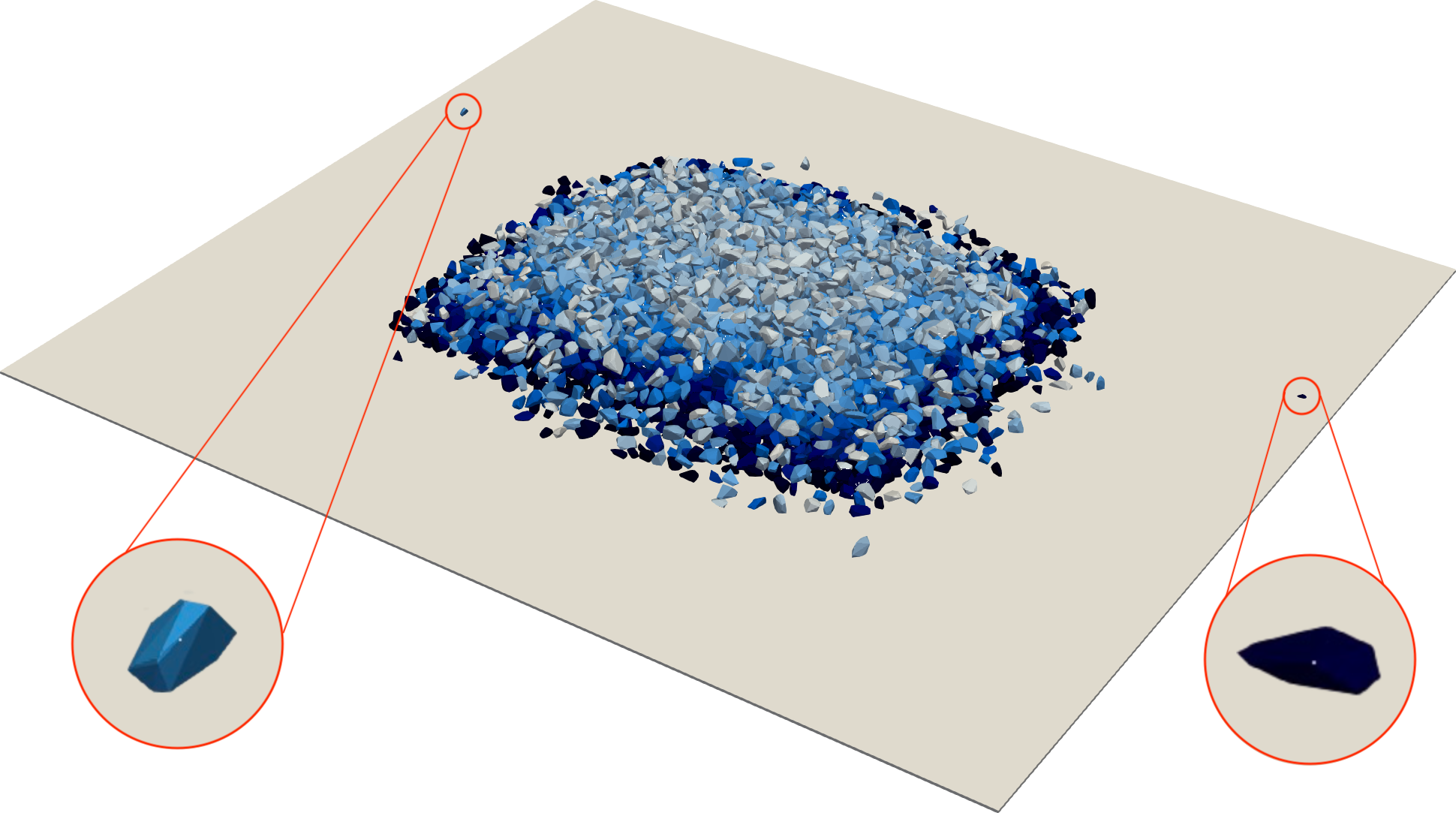}
	\caption{High impact velocities on freefall can lead to some grains being ejected from the sample domain.}
	\label{fig:error_grain.png}
\end{figure}

A concrete example is ballast-track initialization in DEM where grains start in a randomized state (for example, arbitrarily placed in a grid in the air with large gaps between them) and settle into a compacted state (\autoref{fig:fall_first.jpg-fall_final.jpg}). This stage (one that is a very common initalization method) requires very small time steps, sometimes in the microsecond range for ballast-sized particles, to keep contact dynamics stable during impact and settling. For a 1.2-meter track section with approximately 5000-10000 particles, initialization alone can require upto 8 hours on modern hardware. Railway engineers, however, need many such configurations for maintenance, degradation, and dynamic-response studies \citep{aelaDiscreteElementModelling2024}. The approach adopted in this paper is to learn the distributions of the compacted state directly from a few hundred samples that have initialized with DEM, which removes the expensive deposition-and-settling phase for future sample generation, and also enables generation of samples of much larger sizes. This also reduces the chances of grains scattering far away from each other as they fall with high velocities in this traditional initialization method. An example of this unwanted ejection behavior of particles is illustrated in \autoref{fig:error_grain.png}.

\subsection{Contribution}
\label{subsec:contribution}
This paper proposes a 3D diffusion pipeline implementing a generative method based on Markovian stochastic process \citep{hoDenoisingDiffusionProbabilistic2020}. The pipeline operates in two stages. First, an unconditional diffusion model generates independent 3D voxel blocks; second, an inpainting model seamlessly joins these blocks by filling masked boundary regions using context from adjacent blocks. Such a generative strategy mainly leverages initially, a non-guided diffusion models for small sample generation, and subsequently, an adapted 3D inpainting model that renders large granular assemblies of required dimensions by aggregating smaller ones coherently.

The 3D unconditional model uses a standard 2D implementation of diffusion models traditionally applied on images, adapted to 3D occupancy data, while the inpainting model extends this by conditioning on masked boundary regions using a RePaint-style guidance strategy \citep{lugmayrRePaintInpaintingUsing2022}. The architectural details are presented in Section~\ref{sec:methodology} more in depth. In a typical example, a speedup of 100x was empirically measured compared to a similar DEM simulation of a railway track with around 100k particles. This includes the time taken to run a few "settling steps" with DEM to stabilize the assembly configuration.

In summary, the key contribution of this paper pertains the domain-adapted, end-to-end generation pipeline that creates arbitrarily long granular media assemblies of scales previously unattainable. The model presented in this paper ensures a physically realistic generation of granular assemblies with comparison of key granular media properties. Both statistical and granular media specific metrics (packing density, coordination, granulometry, anisotropy, packing fractions, contact force distributions, etc.) are compared to validate the model's performance against DEM baselines. Section~\ref{sec:results} reports quantitative agreement with DEM baselines.

The remainder of this paper is organized as follows. Section~\ref{sec:related_work} reviews diffusion models and existing approaches to granular media generation. Section~\ref{sec:methodology} presents our two-stage pipeline architecture, training procedure, and post-processing steps. Section~\ref{sec:results} validates the approach on railway ballast and lunar regolith, comparing against DEM baselines. Section~\ref{sec:discussion} interprets these results and discusses limitations. Section~\ref{sec:conclusion} concludes with implications for industrial applications.

\section{Related Work}\label{sec:related_work}

\subsection{Diffusion Models}\label{subsec:diffusion_models}
Diffusion models have rapidly emerged as a leading class of generative models thanks to their stability and sample quality. Diffusion models are a class of generative models that learn to reverse a gradual noising process. The idea in short, is to corrupt some given data by adding a known amount of noise and training a network to predict the amount of noise added at each time step. Given training data, noise is iteratively added until the original signal becomes indistinguishable from pure Gaussian noise. A neural network is trained to reverse this process, learning to denoise step by step. At inference time, new samples can be generated by starting from random noise and iteratively denoising. The use and innovation in Diffusion Models have exploded recently. Key contributions span text-to-image and latent diffusion \citep{rombach2022highresolutionimagesynthesislatent}, visual inpainting \citep{wangContextStableVisualConsistentImage2024,lugmayrRePaintInpaintingUsing2022}, and image restoration/super-resolution \citep{moserDiffusionModelsImage2024}. Scalable schedulers and practical tooling have further accelerated adoption \citep{hoDenoisingDiffusionProbabilistic2020,songDenoisingDiffusionImplicit2022,nicholImprovedDenoisingDiffusion2021,liu2022diffusers,peeblesScalableDiffusionModels2023}. The models presented in this paper take architectural hints from many of these papers, many of which have become the standard practice in industry.

The standard forward diffusion process, adopted commonly by these papers, defines a Markov chain that gradually adds Gaussian noise to data $\mathbf{x}_0$ over $T$ time steps, producing increasingly noisy versions $\mathbf{x}_1, \mathbf{x}_2, \ldots, \mathbf{x}_T$. At any time steps $t$, the noisy sample can be computed directly from the original as $\mathbf{x}_t = \sqrt{\bar{\alpha}_t}\mathbf{x}_0 + \sqrt{1-\bar{\alpha}_t}\boldsymbol{\epsilon}$, where $\boldsymbol{\epsilon} \sim \mathcal{N}(0, \mathbf{I})$ and $\bar{\alpha}_t$ is a cumulative product of noise schedule parameters that controls the signal-to-noise ratio at each step \citep{hoDenoisingDiffusionProbabilistic2020}.

The reverse process uses a neural network (typically a UNet architecture \citep{ronnebergerUNetConvolutionalNetworks2015}) to predict the noise $\boldsymbol{\epsilon}$ from the noisy input $\mathbf{x}_t$ and time steps $t$. Training minimizes the mean squared error between predicted and actual noise. During inference, starting from $\mathbf{x}_T \sim \mathcal{N}(0, \mathbf{I})$, the model iteratively removes noise to produce a sample. Various schedulers such as DDPM (Denoising Diffusion Probabilistic Models) \citep{hoDenoisingDiffusionProbabilistic2020} and DDIM (Denoising Diffusion Implicit Models) \citep{songDenoisingDiffusionImplicit2022} define different strategies for this iterative denoising, trading off sample quality against computational cost.

Extending diffusion models from 2D images (the original application of these models as presented in \citep{hoDenoisingDiffusionProbabilistic2020}) to 3D volumetric data presents additional challenges. The cubic growth in memory and computation with resolution, limits direct approaches described in these papers. Latent diffusion methods \citep{rombach2022highresolutionimagesynthesislatent} were introduced to address this by operating in compressed spaces, however, they come at an expense of blurred fine geometric details. For granular media, where sharp particle boundaries and contact geometry are critical, direct voxel diffusion is usually more suitable, as discussed in Appendix~\ref{sec:level_sets}. Following these limitations and the current state of art, the architectural adaptations required for 3D are detailed in Section~\ref{sec:methodology}.

\subsection{DEM Initialization Methods}\label{subsec:dem_init}
Building on the computational challenges outlined in Section~\ref{subsec:motivation_problem_statement}, this subsection reviews existing initialization strategies. There is a broad literature on DEM for ballast and railway applications highlighting computational cost and the difficulty of parallelization \citep{indraratnaCurrentResearchBallasted2017,kleinertModelingLargeScale2013,fredericduboisLMGC902013}. Heuristic initialization refers to rule-based methods that place particles in geometrically admissible positions without full physical simulation; such methods are commonly used but produce only a single admissible configuration per run, if at all. Gravity pluviation and compaction are common strategies that fill up a domain with consecutive particles until a specific void ratio/porosity is achieved. The traditional DEM method of falling particles onto a domain is limited by the small time step required to resolve the increasing kinetic energy \citep{osullivanParticulateDiscreteElement2011}. Other methods like Jodrey-Tory \citep{jodreyComputerSimulationClose1985}, that iteratively remove nearest neighbour overlaps to generate close packing, and Lubachevsky-Stillinger \citep{lubachevskyGeometricPropertiesRandom1990} that grow particle radii while resolving collisions with event-driven dynamics, also exist but suffer from similar limitations. Such strategies can speed up individual test setups but generally do not provide a generative distribution over assemblies in arbitrary physical states. Moreover, inflation and deflation strategies such as the Lubachevsky-Stillinger method are inherently restricted to simple particle geometries and cannot be directly applied to complex grains like ballast, where irregularity in shape and inherent isotropy in grains would be difficult to reproduce. Since the diffusion models are trained on datasets that contain varying physical states and grains of a wide shape range, the trained models are implicitly able to explore such ranges during the generation process, essentially learning the underlying distribution and are able to produce multiple samples of required configurations in single call to the model.

\section{Methodology}\label{sec:methodology}
\begin{figure*}[htbp]
    \centering
    \begin{tikzpicture}[
        data/.style={
            rectangle, rounded corners=2pt, draw=black!70, fill=blue!12,
            minimum width=1.7cm, minimum height=0.75cm, align=center, font=\small
        },
        process/.style={
            rectangle, rounded corners=2pt, draw=black!70, fill=orange!15,
            minimum width=1.7cm, minimum height=0.75cm, align=center, font=\small
        },
        model/.style={
            rectangle, rounded corners=5pt, draw=black!80, fill=green!20,
            minimum width=1.8cm, minimum height=0.8cm, align=center, font=\small\bfseries
        },
        output/.style={
            rectangle, rounded corners=2pt, draw=black!70, fill=purple!15,
            minimum width=1.7cm, minimum height=0.75cm, align=center, font=\small
        },
        imgnode/.style={inner sep=2pt, draw=gray!50, rounded corners=2pt,
            minimum width=3.0cm, minimum height=2.8cm},
        arrow/.style={-Stealth, thick, black!70},
        dashedarrow/.style={-Stealth, thick, black!50, dashed},
        lbl/.style={font=\scriptsize, text=black!60}
    ]

    \def\colA{0}       
    \def\colB{3.4}     
    \def\colC{6.8}     
    \def\colD{10.6}    
    \def\rowI{0}       
    \def\rowII{-1.6}   
    \def\rowIII{-3.2}  
    \def\imgw{2.8cm}
    \def\imgh{2.4cm}

    \node[font=\small\itshape, text=blue!70]         at (\colA, 0.8) {Training};
    \node[font=\small\itshape, text=green!60!black]  at (\colC, 0.8) {Generation};
    \node[font=\small\itshape, text=purple!70]       at (\colD, 0.8) {Output};

    \node[data]    (dem)    at (\colA, \rowI)   {DEM\\Simulations};
    \node[process] (voxel)  at (\colA, \rowII)  {3D\\Voxelization};
    \node[data]    (blocks) at (\colA, \rowIII) {Voxel Blocks\\$32{\times}64{\times}64$};

    \node[model] (uncond)  at (\colB, \rowI)  {Unconditional\\Diffusion};
    \node[model] (inpaint) at (\colB, \rowII) {Inpainting\\Diffusion};

    \node[data]    (genblocks) at (\colC, \rowI)   {Generated\\Blocks};
    \node[process] (stitch)    at (\colC, \rowII)  {Block\\Stitching};
    \node[data]    (long)      at (\colC, \rowIII) {Stitched\\Volume};

    \node[output]  (demout)    at (\colD, \rowI)   {DEM-Ready\\Assembly};
    \node[output]  (grains)    at (\colD, \rowII)  {Grain\\Meshes};
    \node[process] (watershed) at (\colD, \rowIII) {Watershed\\Segmentation};

    \def\imgy{-6.2}
    \node[imgnode] (img_block) at (\colA, \imgy) {%
        \includegraphics[width=\imgw, height=\imgh, keepaspectratio]{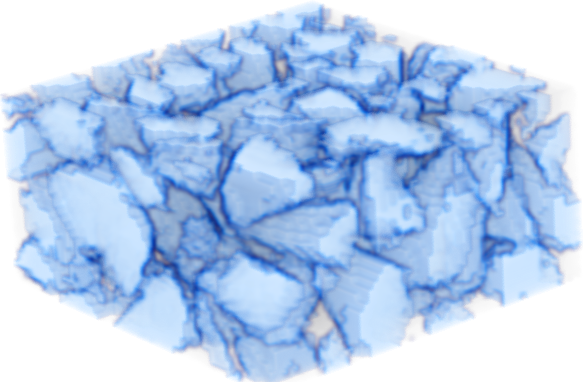}};
    \node[imgnode] (img_unstitch) at (\colB, \imgy) {%
        \includegraphics[width=\imgw, height=\imgh, keepaspectratio]{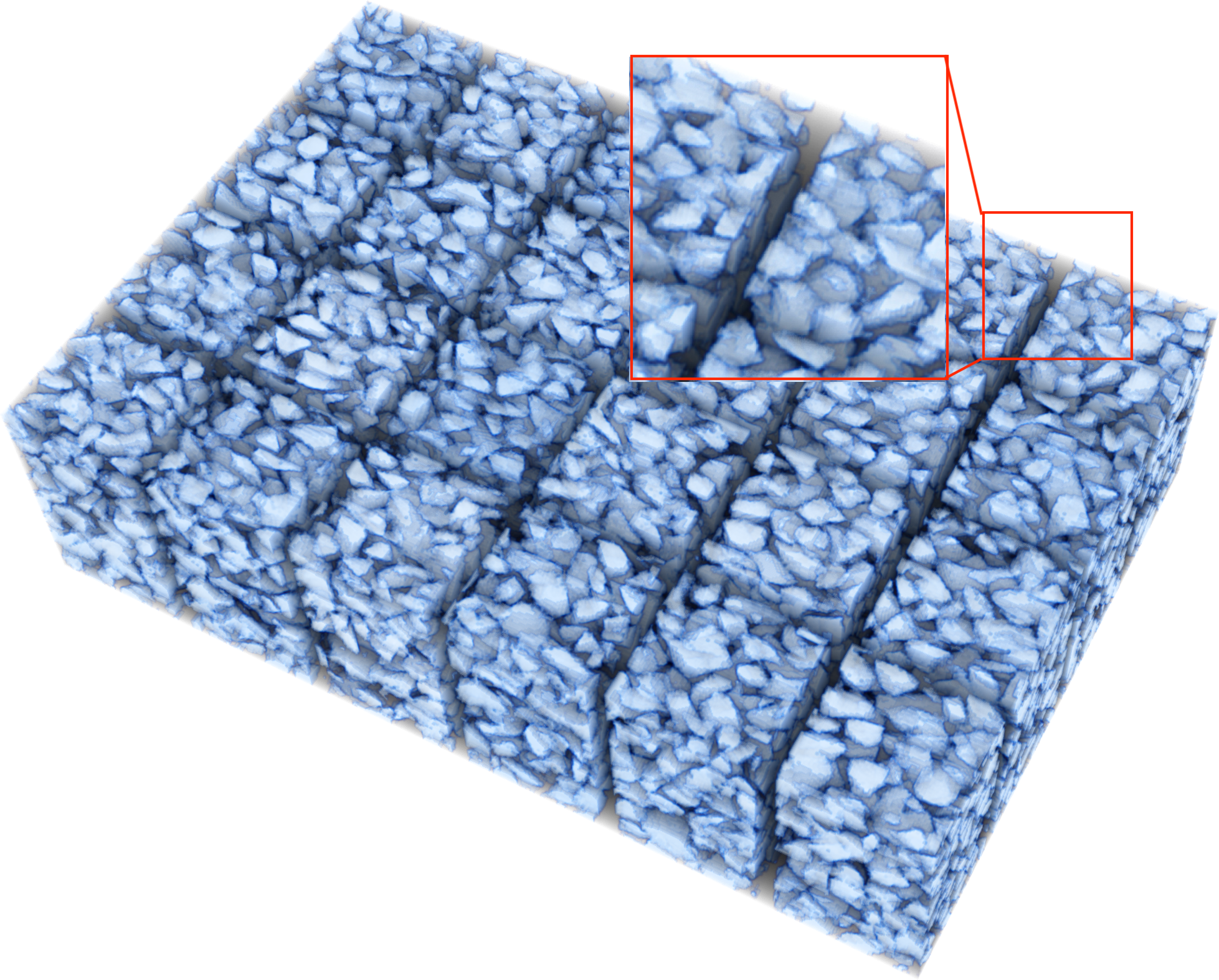}};
    \node[imgnode] (img_stitch) at (\colC, \imgy) {%
        \includegraphics[width=\imgw, height=\imgh, keepaspectratio]{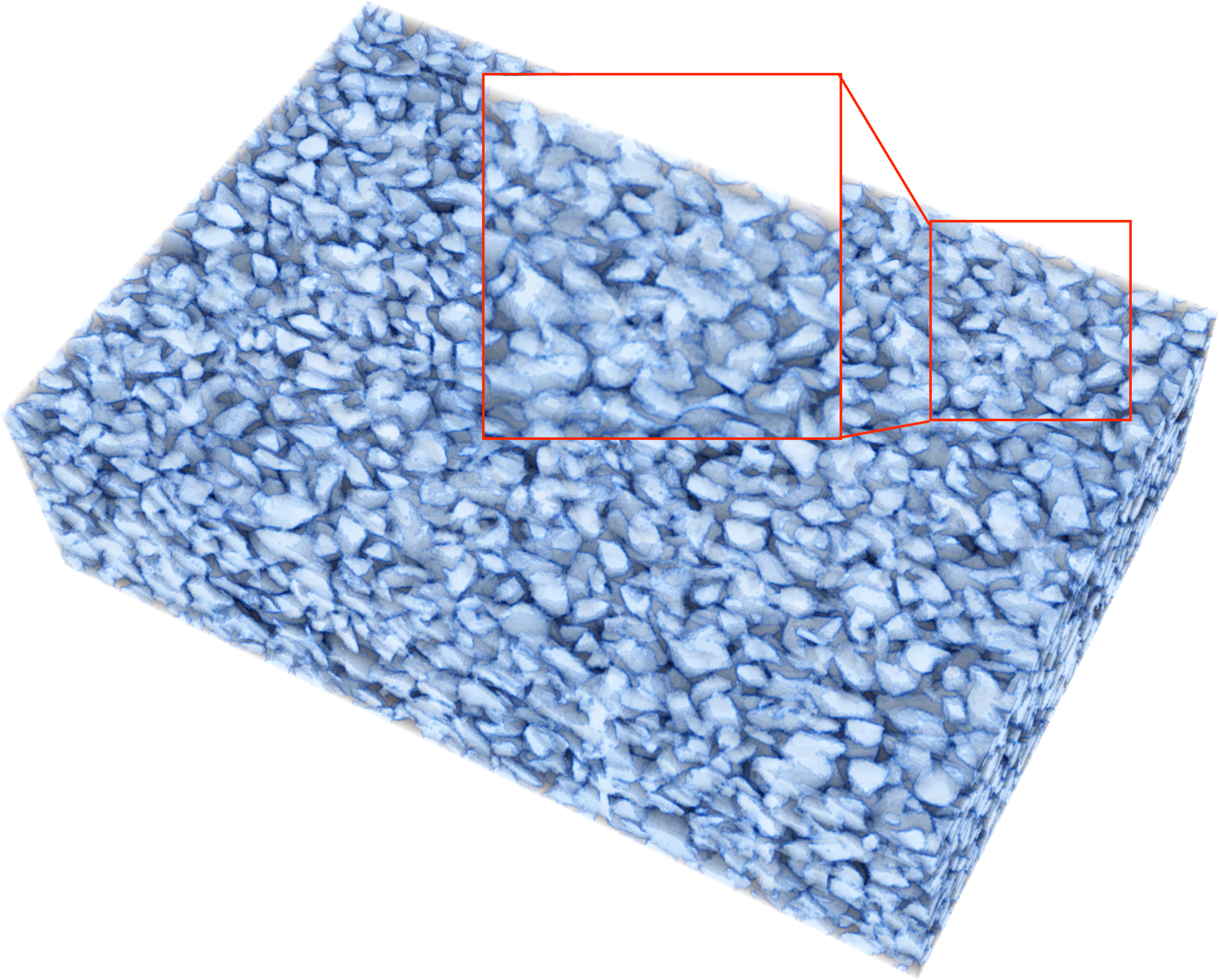}};
    \node[imgnode] (img_seg) at (\colD, \imgy) {%
        \includegraphics[width=\imgw, height=\imgh, keepaspectratio]{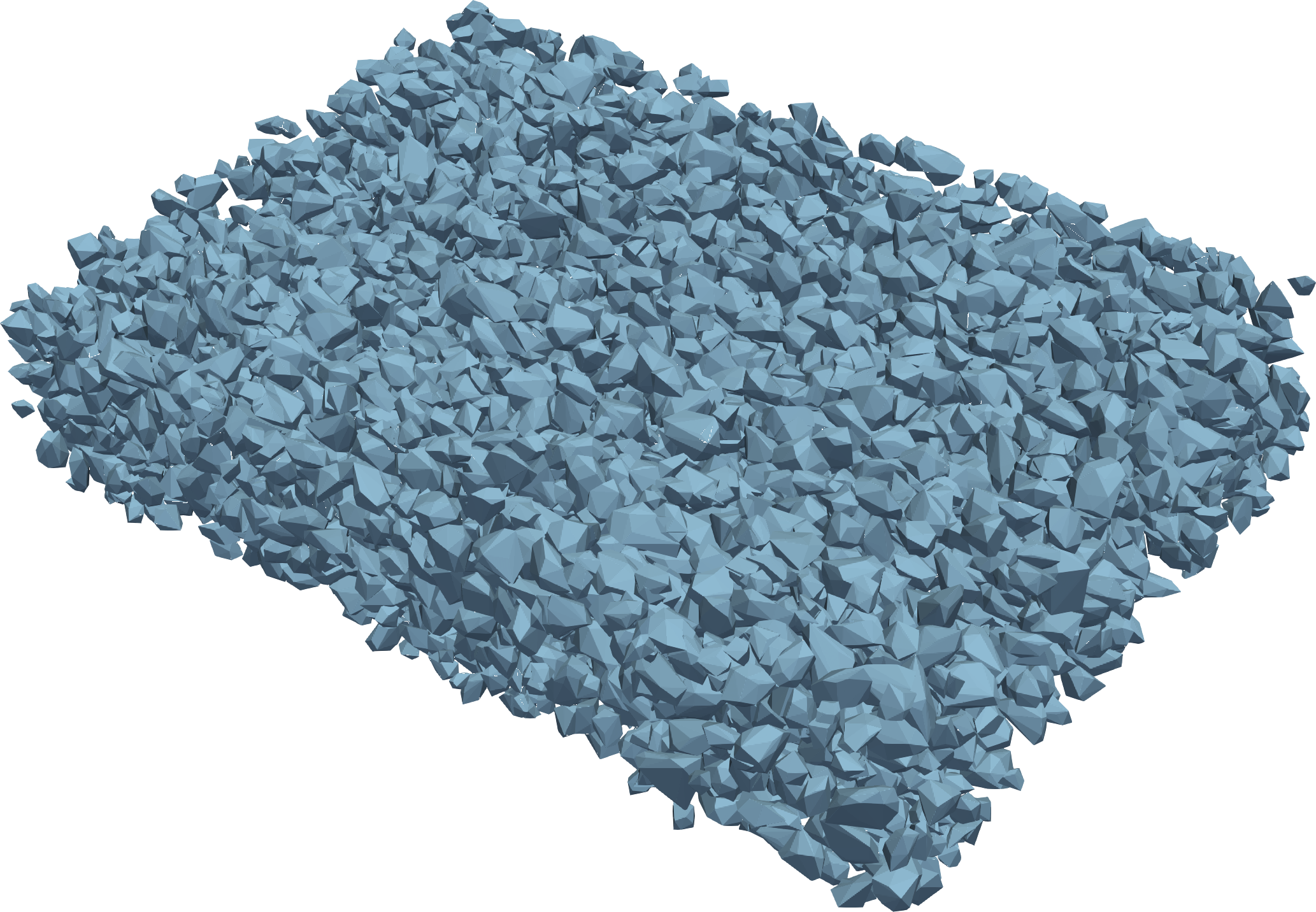}};
    \node[imgnode] (img_lmgc) at (14.2, \imgy) {%
        \includegraphics[width=\imgw, height=\imgh, keepaspectratio]{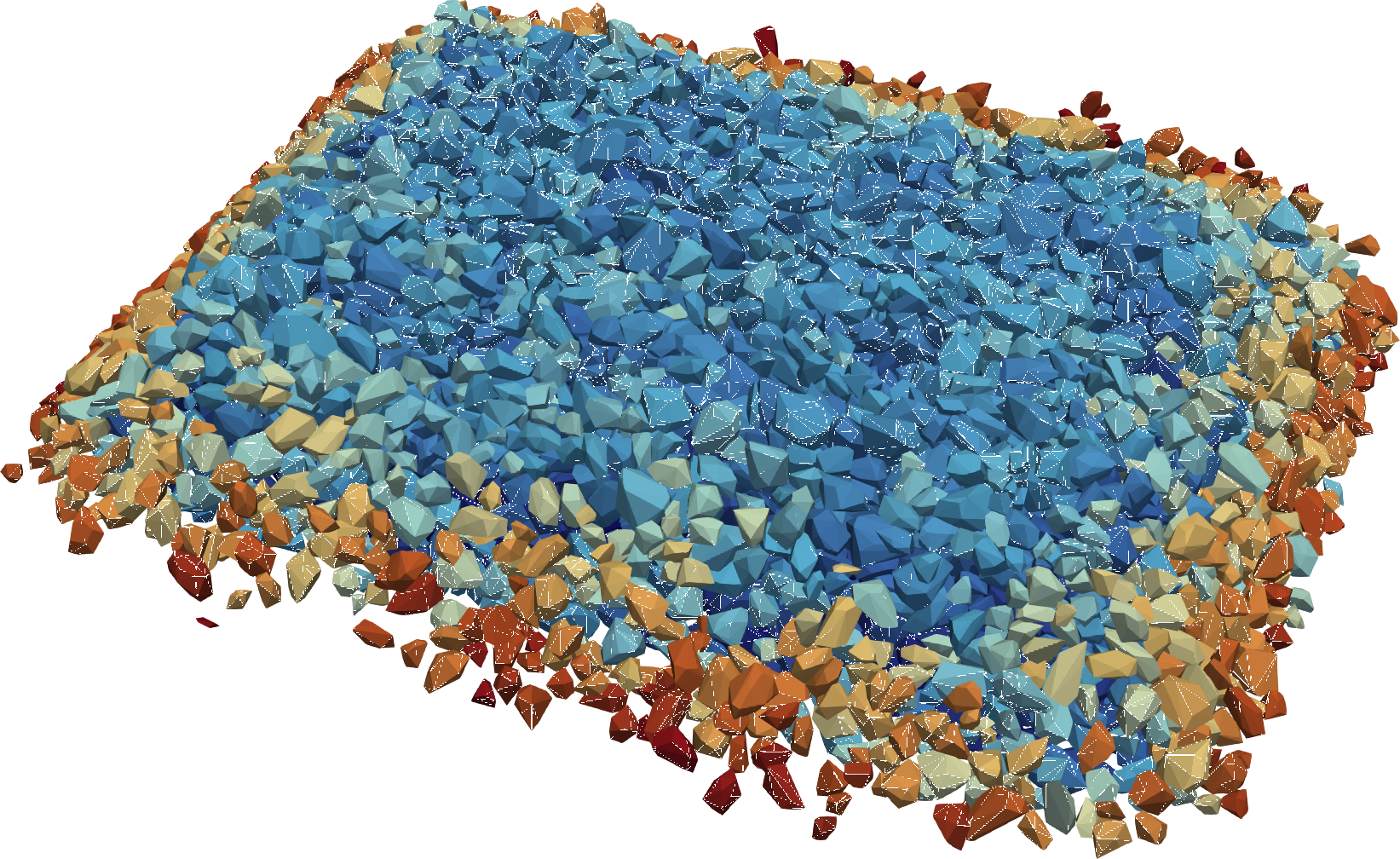}};

    \node[below=0.15cm of img_block,    font=\footnotesize, text=black!70] {(a) Single block};
    \node[below=0.15cm of img_unstitch, font=\footnotesize, text=black!70] {(b) Blocks with gaps};
    \node[below=0.15cm of img_stitch,   font=\footnotesize, text=black!70] {(c) After inpainting};
    \node[below=0.15cm of img_seg,      font=\footnotesize, text=black!70] {(d) Segmented};
    \node[below=0.15cm of img_lmgc,     font=\footnotesize, text=black!70] {(e) LMGC90 simulation};

    \draw[arrow] (dem) -- (voxel);
    \draw[arrow] (voxel) -- (blocks);

    \draw[arrow] (blocks.north east) -- ++(0.65,0) |- (uncond.west)
        node[pos=0.4, left, lbl] {train};
    \draw[dashedarrow] (blocks.east) -- ++(1.15,0) |- (inpaint.west);

    \draw[arrow] (uncond) -- (genblocks) node[midway, above, lbl] {sample};
    \draw[arrow] (genblocks) -- (stitch);
    \draw[arrow] (inpaint) -- (stitch) node[midway, above, lbl] {inpaint};
    \draw[arrow] (stitch) -- (long);

    \draw[arrow] (long) -- (watershed);
    \draw[arrow] (watershed) -- (grains);
    \draw[arrow] (grains) -- (demout);

    \draw[arrow, gray!50, thin] (blocks.south) -- (img_block.north);
    \draw[arrow, gray!50, thin] (inpaint.south) -- (img_unstitch.north);
    \draw[arrow, gray!50, thin] (long.south) -- (img_stitch.north);
    \draw[arrow, gray!50, thin] (watershed.south) -- (img_seg.north);
    \draw[arrow, gray!50, thin] (demout.east) -| (img_lmgc.north)
        node[pos=0.2, above, lbl] {LMGC90};

    \end{tikzpicture}
    \caption{Overview of the diffusion-based granular assembly generation pipeline. Training data from DEM simulations is voxelized into $32{\times}64{\times}64$ voxel grids (a). An unconditional diffusion model generates new blocks, which are placed with gaps (b). An inpainting diffusion model fills the gaps to create seamless stitching (c). Watershed segmentation extracts individual grain meshes (d), which are assembled into DEM-ready geometry and run in LMGC90 for final simulation output (e).}
    \label{fig:pipeline}
\end{figure*}

Having established the theoretical foundations and surveyed existing approaches, the methodology of this paper for scalable granular assembly synthesis is presented. The diffusion-based pipeline for granular media synthesis operates on binary 3D voxel grids. The extraction of these grids will be discussed in \autoref{sub: dataset}. The pipeline proceeds in two stages. Firstly, an unconditional 3D diffusion model generates small, independent voxel blocks representing packed granular media. Following that, an inpainting model seamlessly joins these blocks by conditioning on masked boundary regions. The section first describes the neural network architecture, then the dataset preparation, and finally the large-scale assembly procedure. \autoref{fig:pipeline} displays an overview of the complete pipeline.

\subsection{Neural Network Architecture}
\label{ss: nn archi}
The generative model is based on Denoising Diffusion Probabilistic Models~\citep{hoDenoisingDiffusionProbabilistic2020}, which learn to reverse a gradual noising process applied to training data. Let \(\vx_0\) denote the initial (clean) 3D voxel grid drawn from the training distribution. The forward diffusion process \(q(\vx_t | \vx_{t-1})\), meaning the conditional distribution of the next sample \(\vx_t\) given the previous sample \(\vx_{t-1}\), progressively adds Gaussian noise over \(T\) time steps, and is defined as~\citep{nicholImprovedDenoisingDiffusion2021a,chanTutorialDiffusionModels2024a}:
\begin{align}
    q(\vx_t | \vx_{t-1}) &= \Normal(\vx_t;\; \sqrt{1 - \beta_t}\, \vx_{t-1},\; \beta_t \vI) \\
    q(\vx_t | \vx_0) &= \Normal(\vx_t;\; \sqrt{\bar{\alpha}_t}\, \vx_0,\; (1 - \bar{\alpha}_t) \vI)
\end{align}
where \(\Normal(\,\cdot\,;\,\vmu,\,\vSigma)\) denotes the multivariate Gaussian distribution with mean \(\vmu\) and covariance \(\vSigma\), and \(\vI\) is the identity matrix. The noise schedule \(\{\beta_t\}_{t=1}^T\) is a set of fixed hyperparameters that control the rate of noise addition at each step. The quantities \(\alpha_t = 1 - \beta_t\) and \(\bar{\alpha}_t = \prod_{s=1}^t \alpha_s\) (where each \(\alpha_s = 1 - \beta_s\)) provide a cumulative measure of the signal retained at step \(t\).

The reverse process \(p_\theta(\vx_{t-1} | \vx_t)\) recovers clean data from noise and is parameterized by a neural network with learnable parameters \(\theta\):
\begin{align}
    p_\theta(\vx_{t-1} | \vx_t) &= \Normal(\vx_{t-1};\; \mu_\theta(\vx_t, t),\; \Sigma_\theta(\vx_t, t)) \\
    \mu_\theta(\vx_t, t) &= \frac{1}{\sqrt{\alpha_t}} \left( \vx_t - \frac{\beta_t}{\sqrt{1 - \bar{\alpha}_t}} \epsilon_\theta(\vx_t, t) \right)
    \label{eqn:reverse_process}
\end{align}
where \(\epsilon_\theta(\vx_t, t)\) is the network estimate of the injected noise at step \(t\), and is the primary learned output. The variance \(\Sigma_\theta(\vx_t, t)\) follows a fixed schedule \(\beta_t \vI\) from~\citet{hoDenoisingDiffusionProbabilistic2020}, so it is not learned. In summary, the learnable part is the noise-prediction network parameters (\(\theta\) or \(\phi\) for inpainting), while scheduler coefficients (\(\beta_t, \alpha_t, \bar{\alpha}_t\)) and the number of denoising steps are user-defined hyperparameters.

The architecture consists of a 3D UNet backbone adapted from the standard 2D UNet design~\citep{ronnebergerUNetConvolutionalNetworks2015} commonly used in image diffusion models~\citep{hoDenoisingDiffusionProbabilistic2020,von-platen-etal-2022-diffusers}.

\subsubsection{Unconditional model: 3D UNet backbone and noise scheduling}
\begin{figure*}[htb]
\centering
\includegraphics[width=\textwidth,height=0.45\textheight,keepaspectratio]{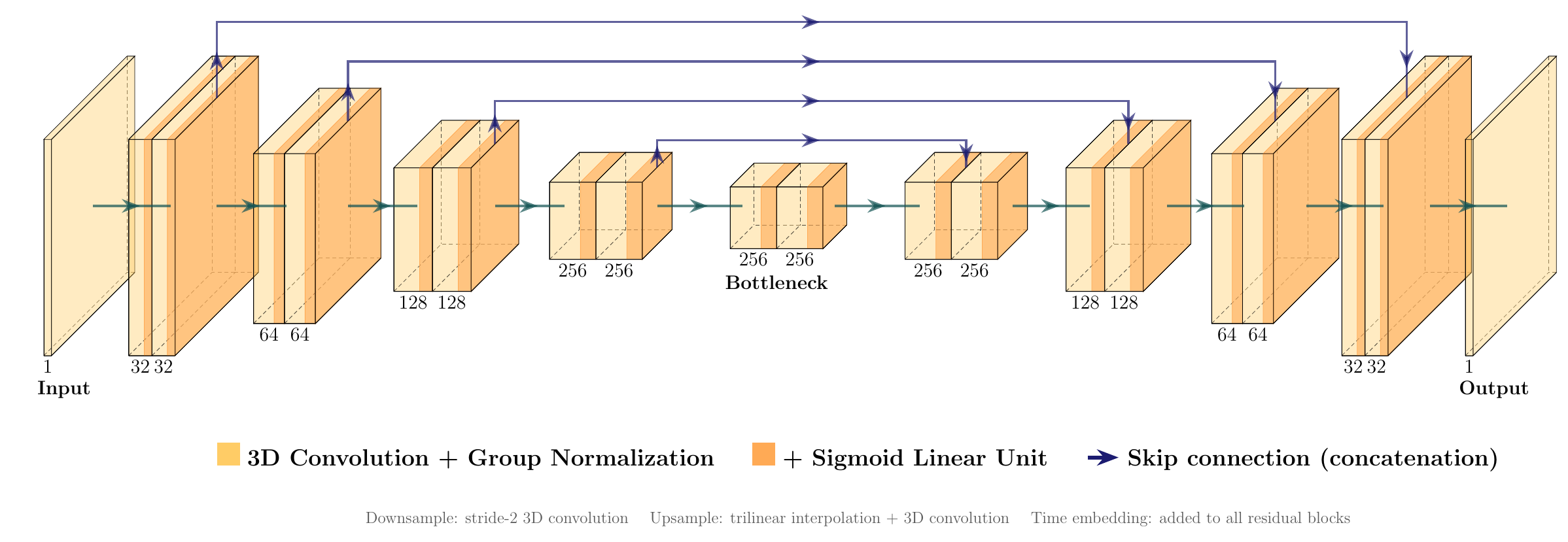}
\caption{3D UNet architecture for voxel-based diffusion. The encoder (left) progressively reduces spatial dimensions via stride-2 3D convolutions while increasing the number of channels ($32{\rightarrow}64{\rightarrow}128{\rightarrow}256$). Each level contains two residual blocks with Group Normalization and Sigmoid Linear Unit activation. The bottleneck (center) processes features at the most coarse resolution and contains the condensed features from the previous layers. The decoder (right) upsamples back to the original resolution via trilinear interpolation followed by convolution. Skip connections (curved arrows) merge encoder features into the corresponding decoder levels (a technique proven to improve the performance of the model). For inpainting, the input layer accepts 3 channels instead of 1 (see Section~\ref{sec:inpainting_extension}).}
\label{fig:unet_architecture}
\end{figure*}

The backbone replaces all 2D convolutions in the standard UNet~\citep{ronnebergerUNetConvolutionalNetworks2015} with their 3D counterparts to operate on volumetric data. The encoder pathway uses residual blocks~\citep{heDeepResidualLearning2016} composed of two successive 3D convolutions, each followed by Group Normalization~\citep{wuGroupNormalization2018} and the Sigmoid Linear Unit (SiLU) activation function~\citep{elfwingSigmoidWeightedLinearUnits2018}. Downsampling is performed via stride-2 convolutions, progressively increasing the channel dimensions through the blocks. The decoder pathway mirrors the encoder, using either transposed convolutions or nearest-neighbor upsampling followed by a convolution to restore spatial resolution. An optional bottleneck layer reduces the number of channels during upsampling, aggregating features from the encoder via skip connections that merge corresponding encoder and decoder feature maps. The noise scheduler follows the standard Denoising Diffusion Probabilistic Model (DDPM)~\citep{hoDenoisingDiffusionProbabilistic2020} or the Denoising Diffusion Implicit Model (DDIM)~\citep{songDenoisingDiffusionImplicit2022}, with a squared cosine noise schedule~\citep{nicholImprovedDenoisingDiffusion2021}. The complete architecture is illustrated in \autoref{fig:unet_architecture}.

The training objective for the unconditional model is to minimize the mean squared error (MSE) between the predicted noise and the true noise:
\begin{align}
  \mathcal{L}_{\mathrm{uncond}} = \E_{t, \vx_0, \vepsilon} \left[ \| \vepsilon - \vepsilon_\theta(\sqrt{\bar{\alpha}_t}\vx_0 + \sqrt{1-\bar{\alpha}_t}\vepsilon, t) \|^2 \right]
  \label{eqn:uncond_loss}
\end{align}
where $\vepsilon_\theta$ denotes the noise prediction network parameterized by $\theta$.

\subsubsection{Extension to inpainting model}\label{sec:inpainting_extension}
Building on the unconditional model, the inpainting model enables seamless assembly of multiple blocks. Simply, the model keeps the boundary voxels that are already known and generates only the missing seam between neighboring blocks. To do this, the network receives three aligned inputs; the noisy voxel grid \(\vx_t\), a binary mask \(\vm\) that marks unknown voxels (1) and known voxels (0), and the known context \(\vx_0^{\mathrm{known}} = \vx_0 \odot (1-\vm)\). These three tensors are stacked along the channel dimension, giving a 3-channel input. The only architectural change is in the first convolutional layer, which now takes a 3 input channels instead of 1. All later layers and the single-channel output remain unchanged.

The network learns to generate content in the masked region that is consistent with the surrounding context, effectively filling in the seam between blocks. The training process follows the same objective as the unconditional model (Eq.~\ref{eqn:uncond_loss}), with a noise prediction network \(\vepsilon_\phi(\vx_t, t, \vx_0^{\mathrm{known}}, \vm)\) parameterized by \(\phi\), and the loss computed only on masked voxels (detailed in Section~\ref{sec:loss_training}).

During inference, the model follows the repaint procedure introduced by \citet{lugmayrRePaintInpaintingUsing2022}. At every denoising step, the known boundary voxels are re-noised to the same time step and reinserted, while the network prediction is used only in masked regions. This repeated composition keeps the seam consistent with both neighboring blocks throughout the full reverse process, not only at the final iteration. The complete procedure is formalized in Algorithm~\ref{alg:repaint}.

\begin{algorithm}[htb]
\caption{RePaint-guided Inpainting Inference}
\label{alg:repaint}
\begin{algorithmic}[1]
\Require Trained inpainting network $\vepsilon_\phi$, known data $\vx_0^{\mathrm{known}}$, mask $\vm$, number of steps $T$
\Ensure Inpainted sample $\vx_0^{\mathrm{final}}$
\State $\vx_T \sim \mathcal{N}(\mathbf{0}, \mathbf{I})$ \Comment{Sample initial noise}
\For{$t = T, T-1, \ldots, 1$}
    \State $\hat{\vepsilon} \gets \vepsilon_\phi([\vx_t, \vm, \vx_0^{\mathrm{known}}], t)$ \Comment{Predict noise}
    \State $\hat{\vx}_0 \gets \frac{1}{\sqrt{\bar{\alpha}_t}}(\vx_t - \sqrt{1-\bar{\alpha}_t}\,\hat{\vepsilon})$ \Comment{Estimate clean sample}
    \State $\tilde{\vx}_{t-1} \gets \text{DDIMStep}(\vx_t, \hat{\vepsilon}, t)$ \Comment{Reverse diffuse}
    \State $\vz \sim \mathcal{N}(\mathbf{0}, \mathbf{I})$
    \State $\vx_{t-1}^{\mathrm{known}} \gets \sqrt{\bar{\alpha}_{t-1}}\vx_0^{\mathrm{known}} + \sqrt{1-\bar{\alpha}_{t-1}}\,\vz$ \Comment{Re-noise known region}
    \State $\vx_{t-1} \gets (1-\vm) \odot \vx_{t-1}^{\mathrm{known}} + \vm \odot \tilde{\vx}_{t-1}$ \Comment{Composite}
\EndFor
\State $\vx_0^{\mathrm{final}} \gets (1-\vm) \odot \vx_0^{\mathrm{known}} + \vm \odot \hat{\vx}_0$ \Comment{Final composition}
\State \Return $\vx_0^{\mathrm{final}}$
\end{algorithmic}
\end{algorithm}

\subsubsection{Loss function and training parameters}\label{sec:loss_training}
The unconditional training objective was defined in Eq.~\ref{eqn:uncond_loss}. For the inpainting model, a binary mask $\vm$ conditions the model on the known regions $\vx_0^{\text{known}} = \vx_0 \odot (1-\vm)$. The inpainting loss becomes:
\begin{equation}
  \mathcal{L}_{\text{inpaint}}
  =
  \E_{t,\,\vx_{0},\,\vepsilon,\,\vm}
  \left[
  \left\|
  \vm \odot
  \bigl(
  \vepsilon - \vepsilon_{\phi}(\vx_{t},t,\vx_{0}^{\text{known}},\vm)
  \bigr)
  \right\|_{2}^{2}
  \right],
\end{equation}
where $\vepsilon_\phi$ is the inpainting network parameterized by $\phi$. The mask $\vm$ ensures that the model predicts noise only in the masked (unknown) regions, while the known regions provide context.

If weighted loss is enabled, the objective becomes:
\begin{equation}
  \mathcal{L}_{\text{weighted}} =
  \frac{1}{N}
  \sum_{i=1}^{N}
  w_i \bigl(\epsilon_i - \hat{\epsilon}_i\bigr)^2,
  \qquad
  w_i =
  \begin{cases}
  \lambda, & \text{if } m_i = 1,\\[4pt]
  1, & \text{if } m_i = 0,
  \end{cases}
\end{equation}

\(\lambda \) is the masked loss weight (default 2.0). The value of the loss weight is chosen purely based on the empirical observations during the training process, and usually carries no theoretical significance.

 Both models follow the channel blocks configured as \([32, 64, 128, 256]\) with 2 convolution layers  per block. An optional self-attention mechanism~\citep{vaswaniAttentionAllYou2023} was also tested; however, with no significant gain in quality of the samples in the ballast examples, only 3D convolutions were kept, which reduces the computational cost by a significant amount. AdamW optimizer \citep{loshchilovDecoupledWeightDecay2019} is used for both with a starting learning rate of \(10^{-4}\) and a batch size of 32. The learning rate usually is not linear and is chosen based on the empirical observations during the training process, which in the current case, was a linear warmup for the first 5 epochs. The models are trained for around 160 epochs on 2 NVIDIA H100 GPUs with a total of 160 GB of memory. Both models roughly ended up being around 600 Million parameters, which remains on the smaller side for 3D models.

 \subsection{Dataset Construction}
 \label{sub: dataset}
 Training dataset is created by dividing and voxelizing the DEM simulations into Representative Volume Elements (RVEs) with a subset of grains per grid. The number of grains per grid were chosen on the basis of the resolution of the vertices, as to not degrade the final locally convex geometry of the grains too much. The study covers two distinct materials, railway ballast and lunar regolith simulants, with fundamentally different scales and packing behaviors. Each material uses its own independently initialized model pair (unconditional and inpainting). No transfer learning was applied between ballast and regolith, and each pair was trained from scratch on its own dataset. Transfer learning could be explored, however owing to the fact that the models are relatively small and training times are affordable, a dedicated model might be worth the training effort.

 \subsubsection{Railway Ballast Dataset}\label{sec:ballast_dataset}
\begin{figure}[htb]
   \centering
   \begin{subfigure}[htb]{0.45\textwidth}
     \includegraphics[width=\textwidth]{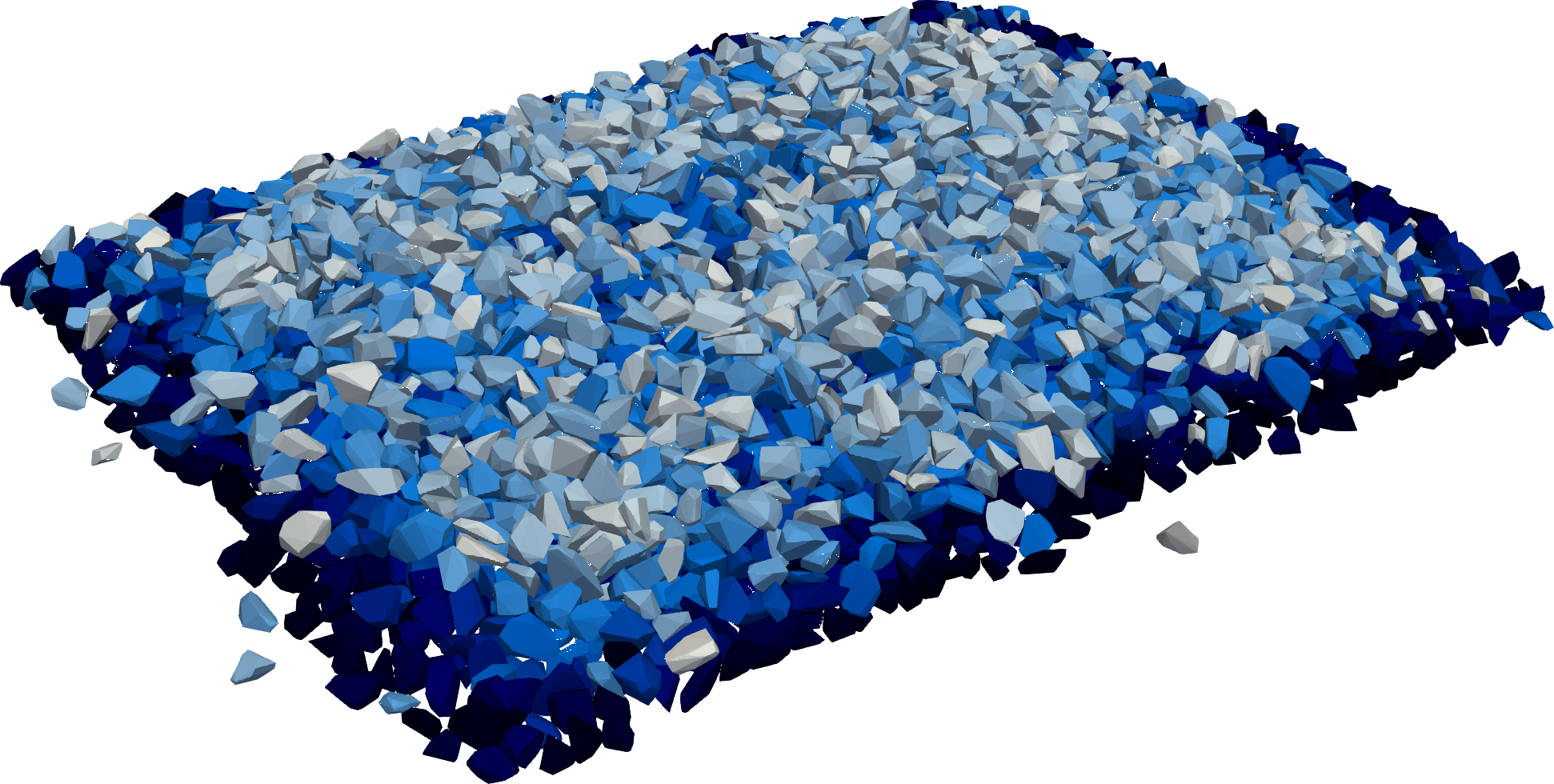}
     \caption{Final state of a ballast track simulation}
     \label{fig: final ballast state}
   \end{subfigure}
   \begin{subfigure}[htb]{0.45\textwidth}
     \includegraphics[width=\textwidth]{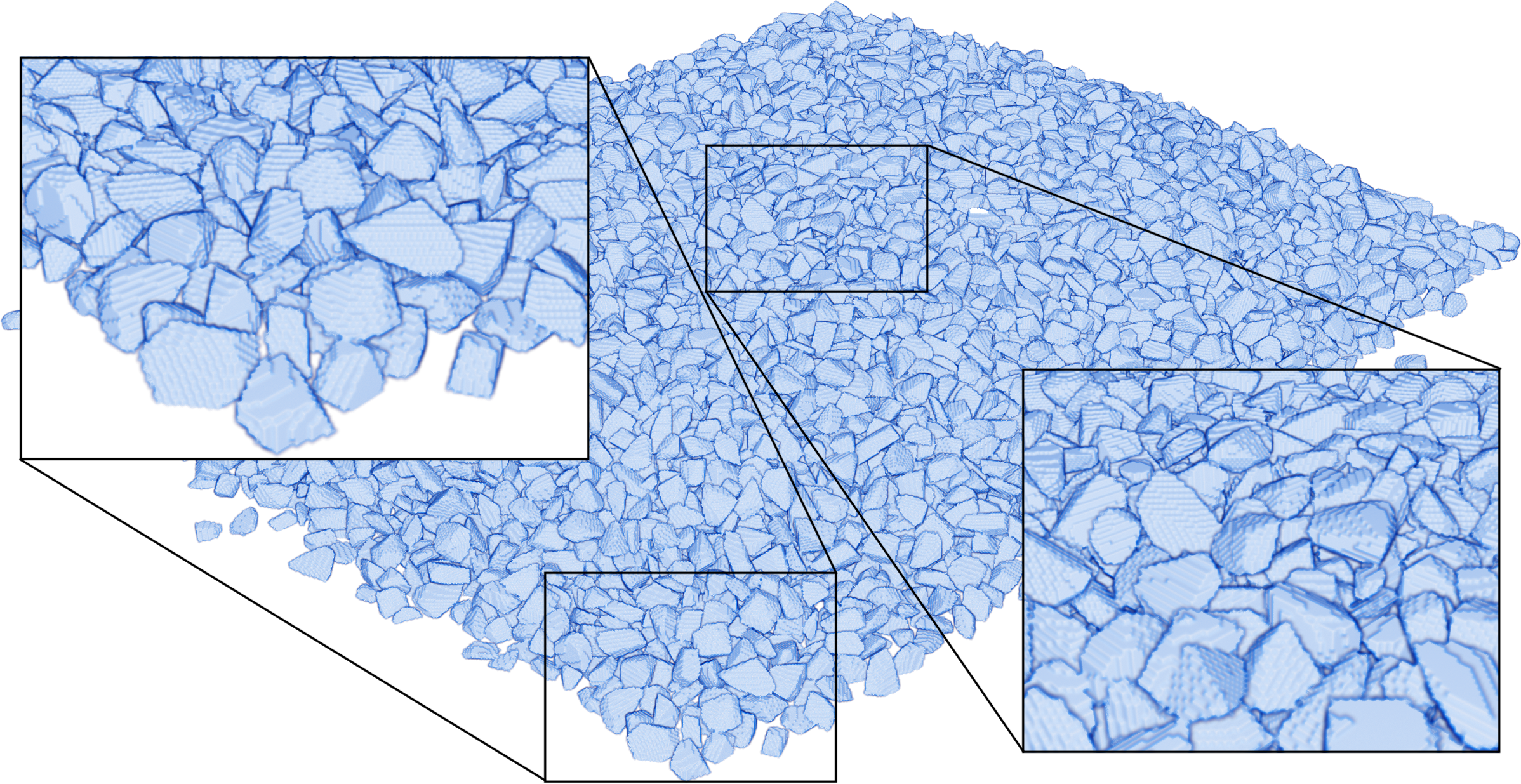}
     \caption{Voxelized grid}
     \label{fig: final ballast state voxelized}

   \end{subfigure}
   \caption{The dataset is procured from DEM simulations of small ballasted tracks in LMGC90 \citep{fredericduboisLMGC902013}. a): a sample of initialized ballast layer, as it comes out of LMGC90. b): the same dataset voxelized into binary occupancy grids.}\label{fig:dim_ballast.png}

 \end{figure}

 The initial database was procured from simulations run for ballasted railway tracks with the LMGC90 code \citep[][software under GNU General Public License (GPL) / CeCILL (GPL-compatible)]{fredericduboisLMGC902013}. Traditionally, a ballasted railway track has a dimension of around 1.435 m of guage width, with ballast flowing across the rails (total width taken as 2.0 m), 0.3 m of height and 0.6 m sleeper to sleeper length in case of SNCF's TGV tracks \citep{decTimeDomainSimulations2022}. The simulations were run for an equivalent of two consecutives sleeper-to-sleeper distances. Hence, the dimensions of one instance of the ballasted track was nearly (1.435 m $\times$ 1.2 m $\times$ 0.3 m), with each ballast particle having a diameter between 25 mm and 50 mm \citep{correaLargeScaleNumericala}. 531 such simulations were run where grain centers are generated in the air and grains are placed in a non-overlapping way on this grid, and allowed to fall under gravity. After a compaction stage, the final states of all the instances were exported as meshes, which contained geometric and topological information on the grains and the assembly. The final state of one instance is shown in \autoref{fig: final ballast state}.

 Mesh files from DEM simulations are processed to extract 2D slices along the $z$-axis (the vertical axis), cut at uniform intervals \citep{vtkBook,van2014scikit}.\label{sec:dataset_slicing} The 2D binary masks are then stacked to create a 3D voxelized representation of the original mesh. This mesh is then cut into smaller non-overlapping volumes of size \(32 \times 64 \times 64\) (a length of around 5 ballast particles across one dimension, corresponding to 120-150 mm), ensuring compatibility with GPU memory constraints and batch processing. An example of the voxelized dataset is shown in \autoref{fig: final ballast state voxelized}.

 The total training hours for the final models explained in \autoref{ss: nn archi} were 12 hours and roughly 16 hours respectively for the unconditional and the inpainting models with these 531 ballast samples. Total number of samples for the final dataset was 30k samples of voxelized data, with a \(80:20\) train test split. The inpainting models took slightly longer at adjusting weights even though both models seemed to converge at around 12 hours mark.

 \subsubsection{Lunar Regolith Simulant Dataset}\label{sec:lunar_dataset}
\begin{figure}[htb]
   \centering
   \begin{subfigure}[htb]{0.49\textwidth}
     \includegraphics[width=\textwidth]{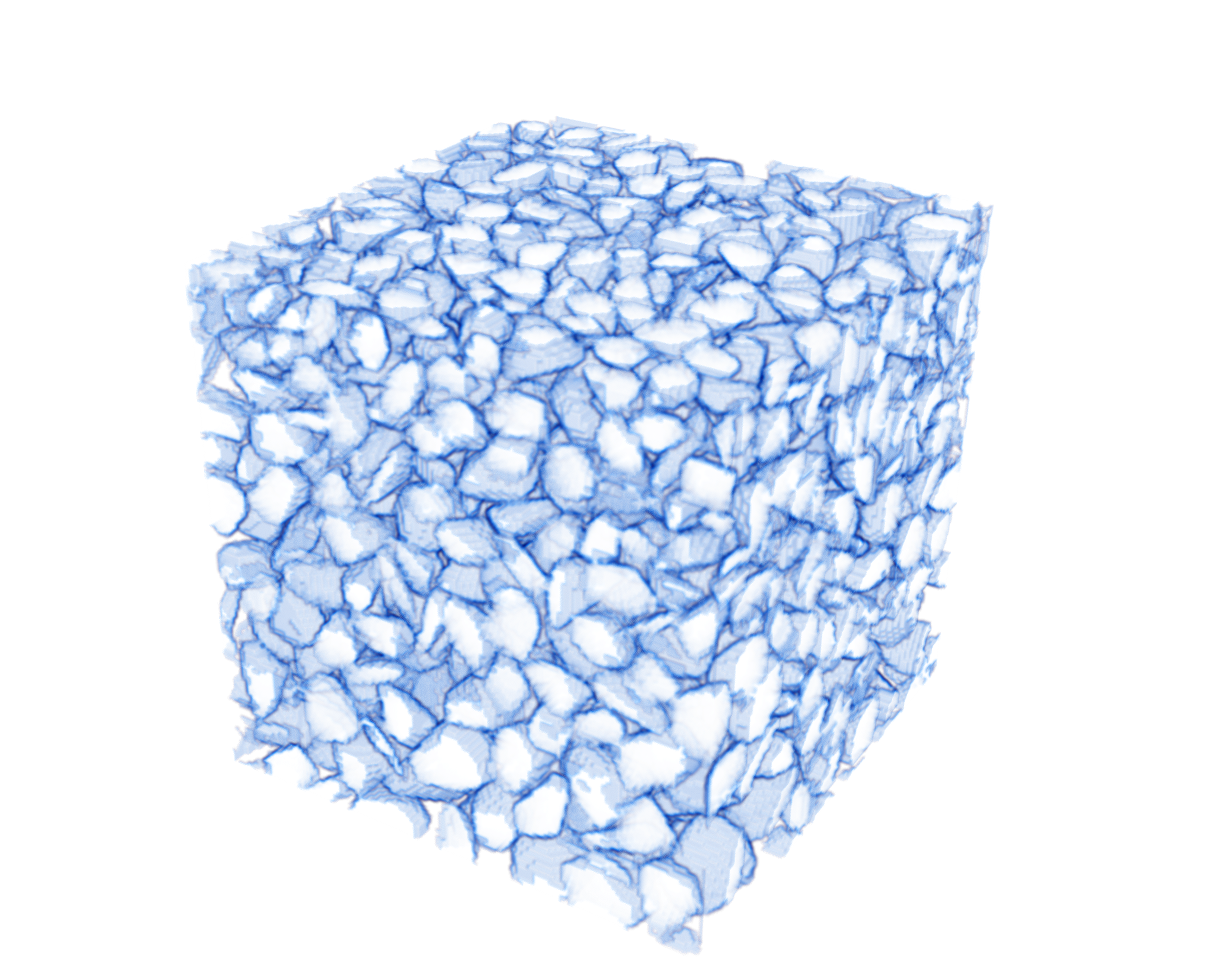}
     \caption{Dataset sample from DEM simulation}
     \label{fig:original_sand_block.png}
   \end{subfigure}
   \begin{subfigure}[htb]{0.49\textwidth}
     \includegraphics[width=\textwidth]{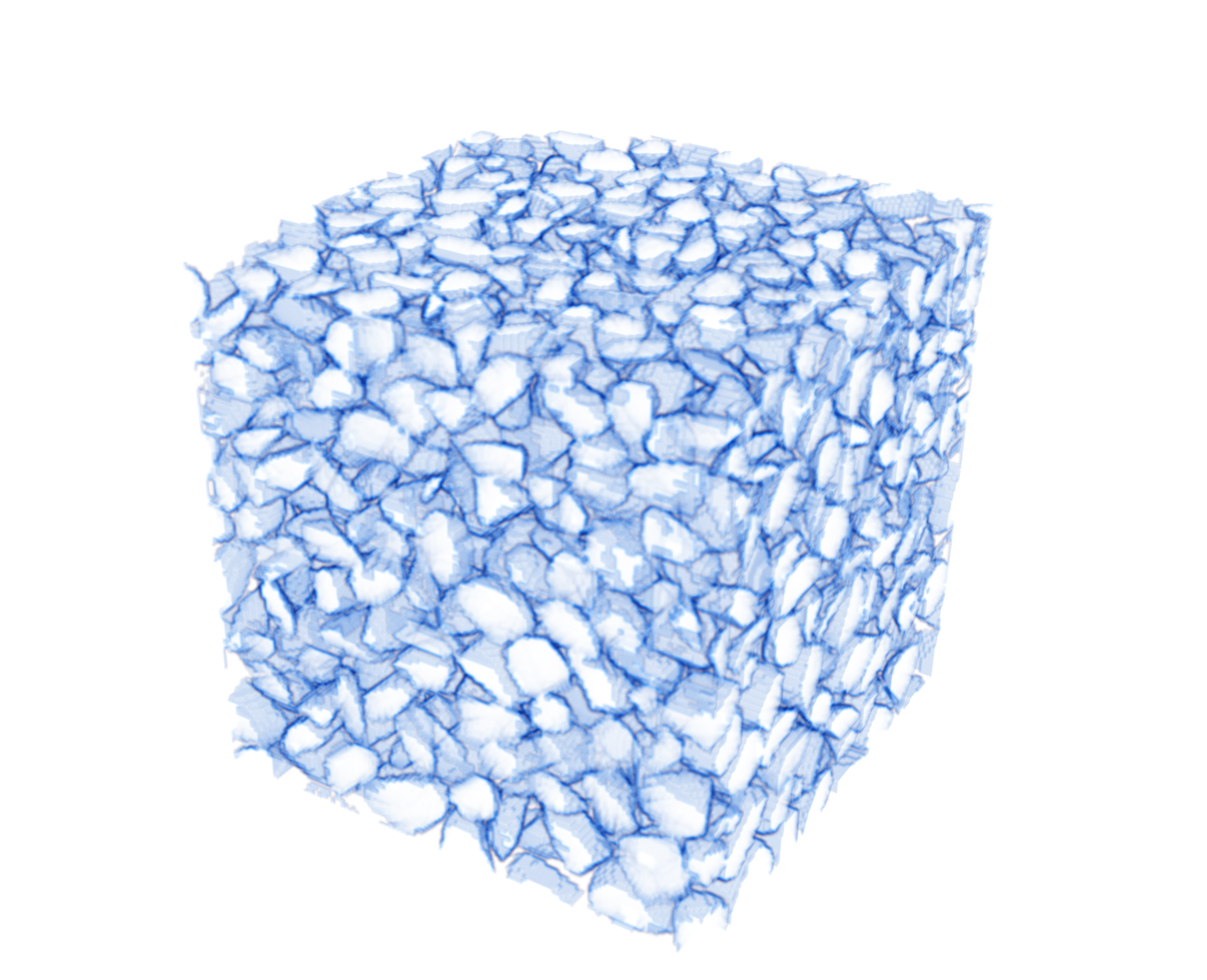}
     \caption{Generated sample from diffusion model}
     \label{fig:ddpm_sand_block.png}
   \end{subfigure}
   \caption{Comparison of (a) original and (b) generated lunar regolith simulant blocks compacted under gravity. The DEM sample took around 1 hour to generate.}
   \label{fig:lunar_blocks}
 \end{figure}

 The second dataset was constructed from synthetic DEM simulations based on the \textbf{NU-LHT-4M} lunar regolith simulant \citep{kafkaThreeDimensionalShape2024}. Unlike the coarse railway ballast (25--50 mm diameter), lunar regolith simulants are fine (sub-millimeter), more cohesive, and exhibit different packing behavior. A library of 2500 morphologically unique grains was created from simplified STL meshes (80--200 faces each), and 1500 DEM simulations were run to generate packed RVEs with packing fractions around 0.5. The voxelization and data processing followed the identical pipeline as for railway ballast, producing $32 \times 64 \times 64$ voxel blocks. Training used the same hyperparameters (Adam optimizer, learning rate as \(10^{-4}\) , batch size as 32 etc) and required approximately 5 hours per model on 2 NVIDIA H100 GPUs. An example voxelized block is shown in \autoref{fig:original_sand_block.png}.

\subsection{The Generation Pipeline}\label{sec:generation_pipeline}

New samples are generated by reversing the diffusion process using the trained UNet models. The inference employs the Denoising Diffusion Implicit Model (DDIM) scheduler~\citep{songDenoisingDiffusionImplicit2022}, although the Denoising Diffusion Probabilistic Model (DDPM) scheduler~\citep{hoDenoisingDiffusionProbabilistic2020} can also be used. For unconditional generation, the process follows the standard reverse diffusion (Eq.~\ref{eqn:reverse_process}). For the inpainting model, the inference process additionally requires a binary mask \(\vm\) and the corresponding known voxel data \(\vx_0^{\text{known}} = \vx_0 \odot (1-\vm)\), which may come from two adjacent samples generated by the unconditional model.

\begin{figure}[htp]
  \centering
  \includegraphics[width=0.60\textwidth]{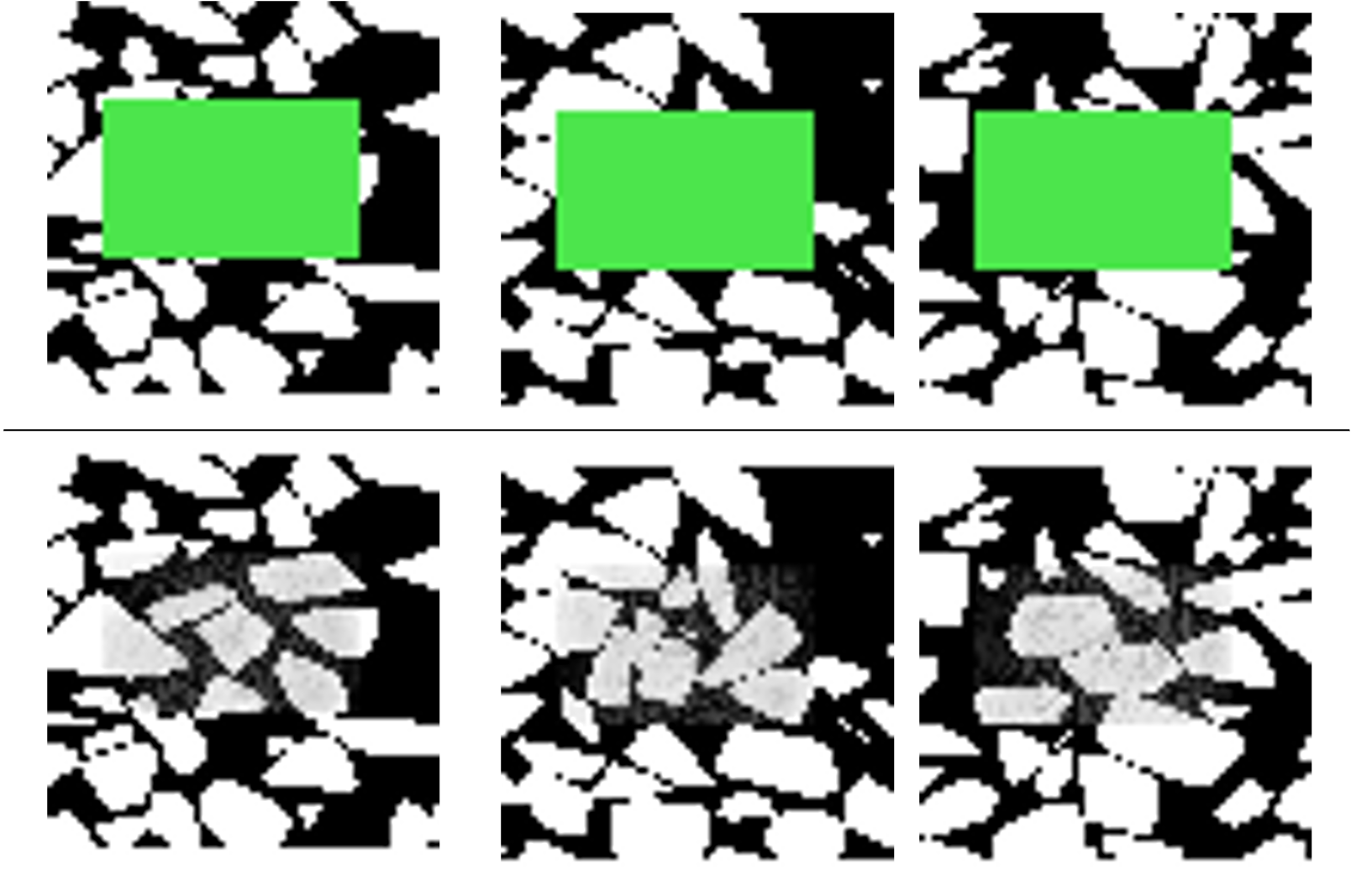}
  \caption{2D slices of 3D voxel grids with missing blocks throughout the vertical axis. The inpainting model predicts the masked (green) portions of the voxel grids (bottom row). Seam regions show Z1 difference $<0.6$ compared to interior regions.}
  \label{fig:Inpainting.png}
\end{figure}

At each denoising step \(t\), the inpainting UNet \(\epsilon_\phi\) receives the merged input \([\vx_t, \vm, \vx_0^{\text{known}}]\) along the channel dimension and predicts the noise \(\hat{\vepsilon}\). The DDIM scheduler then computes the candidate previous sample \(\tilde{\vx}_{t-1}\). Simultaneously, the original known data \(\vx_0^{\text{known}}\) is re-noised to time step \(t{-}1\) via the forward process schedule, yielding \(\vx_{t-1}^{\text{known}}\). The two are composited using the mask: \(\vx_{t-1} = \vm \odot \tilde{\vx}_{t-1} + (1-\vm) \odot \vx_{t-1}^{\text{known}}\). At the final step (\(t{=}1 \to t{=}0\)), the original known data is used directly: \(\vx_0 = \vm \odot \hat{\vx}_0 + (1-\vm) \odot \vx_0^{\text{known}}\). This is the same repaint procedure formalized in Algorithm~\ref{alg:repaint}. An illustration of the inpainting model filling in missing blocks is shown in \autoref{fig:Inpainting.png}.

With both models trained, the generation pipeline proceeds in two stages. First, the unconditional model generates independent voxel blocks; then, the inpainting model joins adjacent blocks by filling masked seam regions. This two-stage approach enables generation of arbitrarily large assemblies while maintaining local statistical consistency.

\subsubsection{Stage 1: Unconditional block generation}\label{sec:unconditional_generation}
\begin{figure}[htb]
  \centering
    \includegraphics[width=0.40\textwidth]{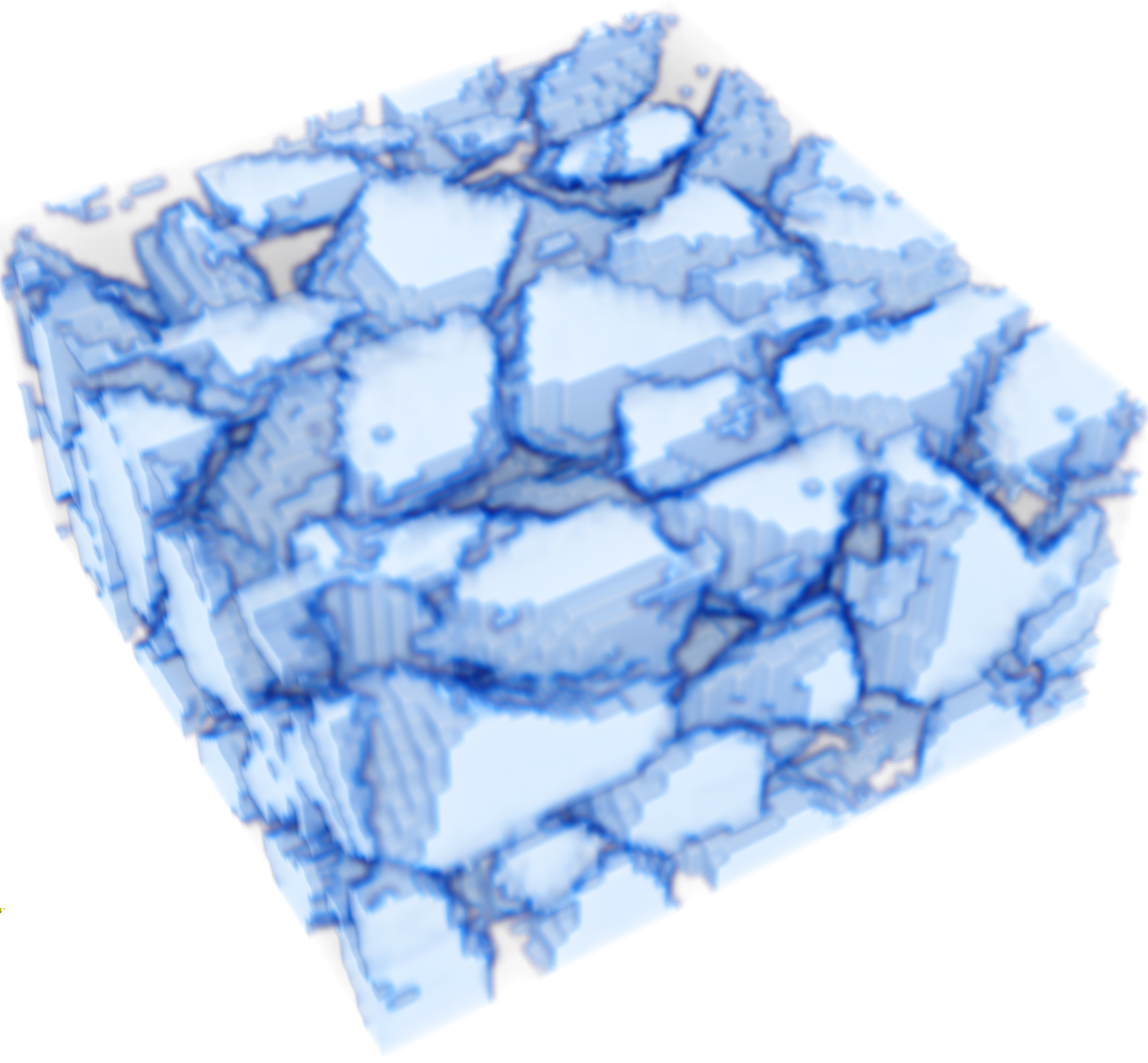}
  \caption{Example of an unconditionally generated voxel grid of ballast grains. The model reproduces sharp contacts and realistic packing (packing fraction within 1.7\% of DEM baseline) in local neighborhoods.}
  \label{fig:uncond_ballast.png}
\end{figure}

The generation for the independent voxel grids follows the DDIM strategy described above. The number of steps for the inference was set to 50, and the number of generated voxel grids was set to 8 in one batch. Generation proceeds in batches to accommodate GPU memory constraints. An example is shown in the \autoref{fig:uncond_ballast.png}.

\subsubsection{Stage 2: Block assembly via inpainting}

For creating larger samples, \textit{N} voxel grids are joined together in a coherent fashion using the inpainting model. As can be seen in \autoref{fig:uncond_ballast.png}, the voxel grid is unconditionally generated and would not make coherent boundary with other unconditionally generated voxel grids when simply joined, without any blending or inpainting technique. The discontinuities at the boundaries would create segmentation issues and would create grains that do not respect the statistical characteristics of the original dataset. The assembly strategy is illustrated in \autoref{fig:stitching_schematic}. Independently generated blocks are placed with a gap between them, and the inpainting model fills in the gap conditioned on the known context from both adjacent blocks, producing seamless transitions. For \(N\) blocks, the resulting assembly has a size of \(N \times Block-Depth + (N{-}1) \times Gap-Depth\) along the stitching direction. Note that this assembly process assumes isotropy across the voxel grids. While gravity during DEM simulation induces slight anisotropy in the vertical direction, the effect remains small for the block sizes considered here and is captured in the fabric tensor statistics reported in Section~\ref{sec:results}.

\begin{figure}[htp]
  \centering
  \includegraphics[width=0.84\textwidth]{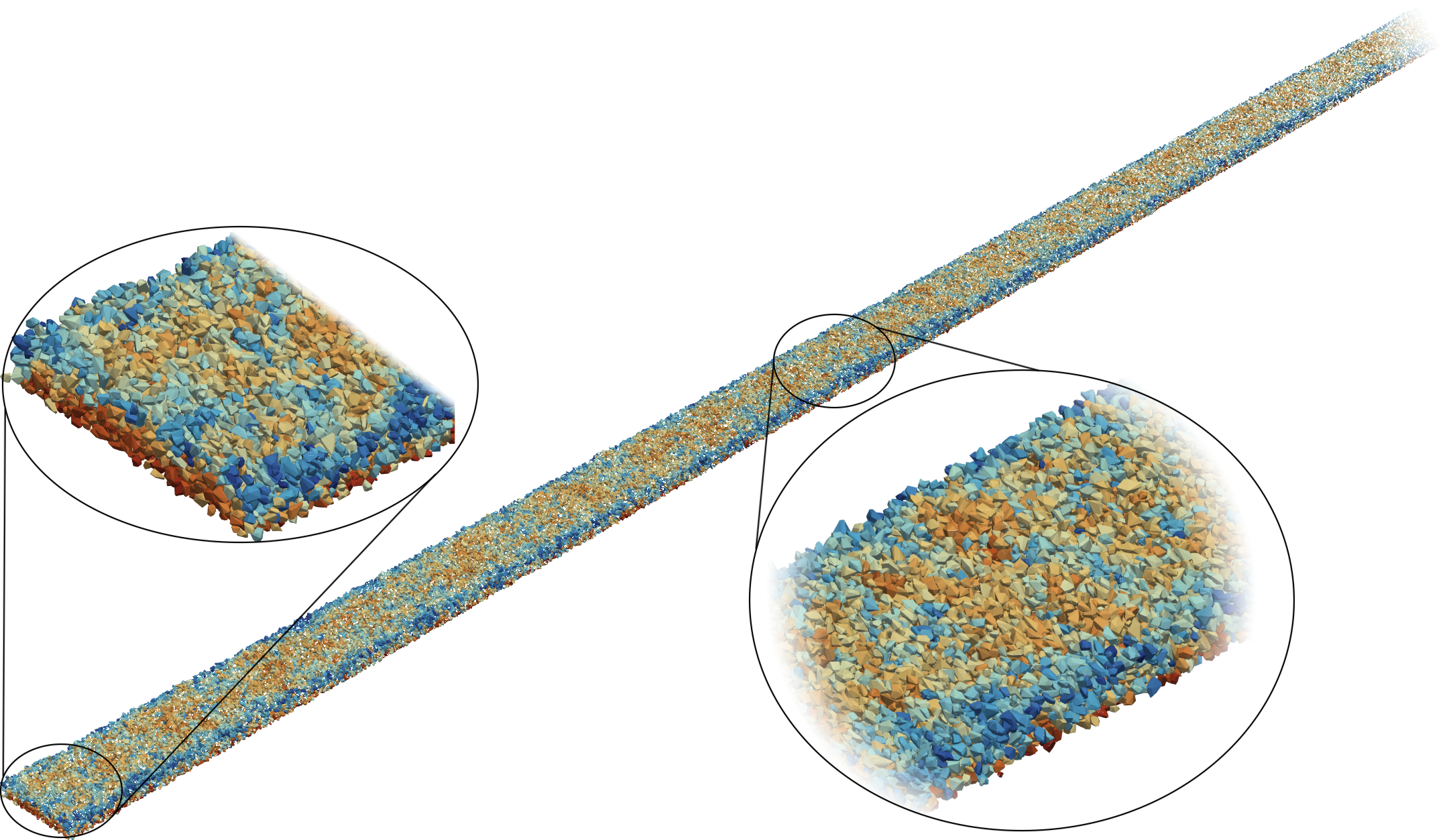}
  \caption{A 100\,m long ballast sample ($\sim$200k particles) generated using the pipeline. Voxel generation took under 60 minutes; the full pipeline (generation + segmentation) took roughly 3 hours on 1 NVIDIA H100 GPU and 8 cpu cores for segmentation and DEM, versus more than 130 hours for an equivalent DEM initialization. A more detailed speed comparison is presented in \autoref{fig:DEMvsDDPMspeedup.png}}
  \label{fig:ballast_long.png}
\end{figure}

\begin{figure}[htb]
  \centering
  \begin{subfigure}[t]{0.48\textwidth}
    \centering
    \includegraphics[width=\textwidth]{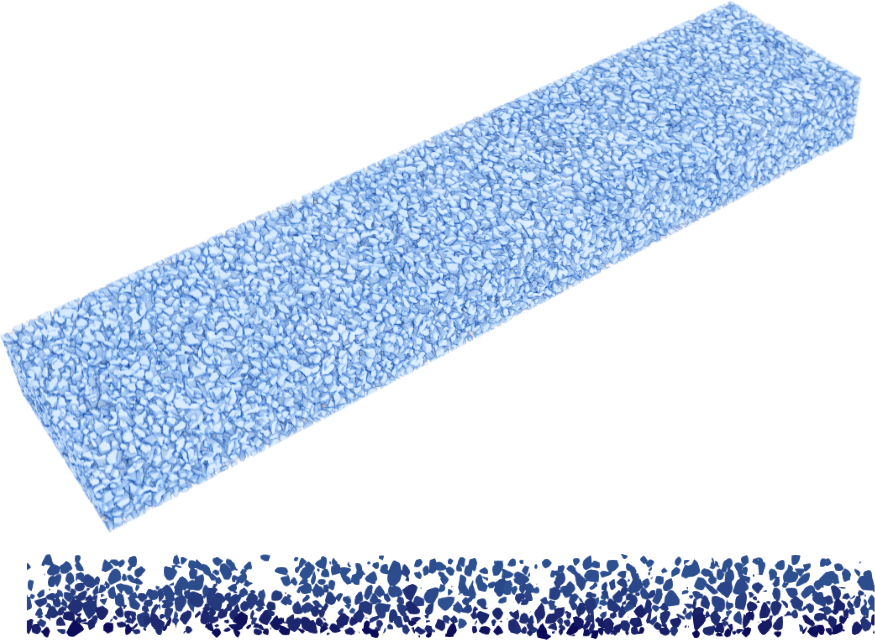}
    \caption{Generated stitched assembly. Top: 3D rendering; bottom: 2D cross-section.}
    \label{sf1:extra_results}
  \end{subfigure}\hfill
  \begin{subfigure}[t]{0.48\textwidth}
    \centering
    \includegraphics[width=\textwidth]{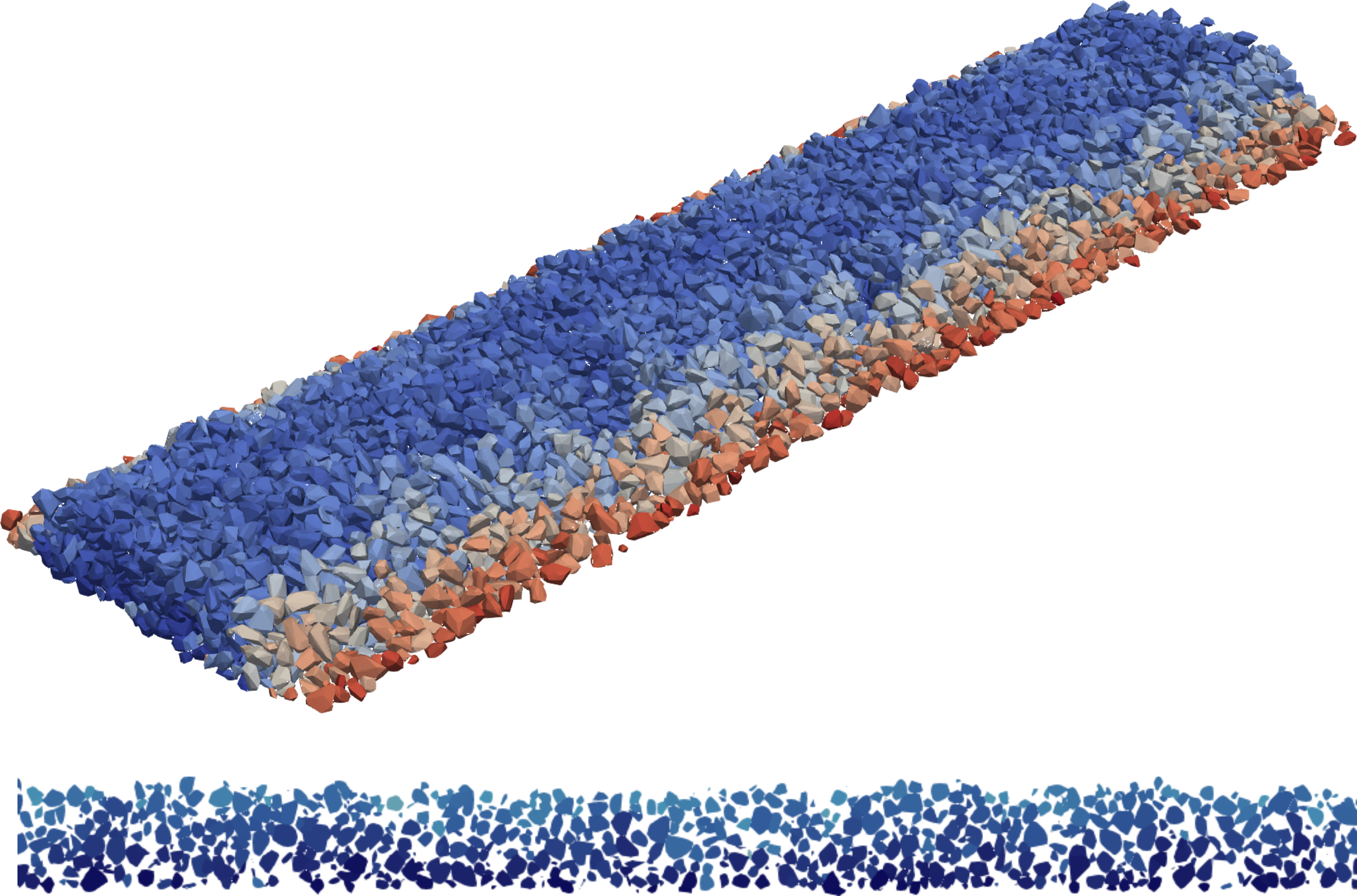}
    \caption{Segmented assembly. Top: 3D rendering colored by normalized height $z$; bottom: 2D cross-section.}
  \end{subfigure}
  \caption{Example stitched volume ($\sim$30k grains): generated assembly (a) and segmented output (b). The colormap in the right panel is used to show relative displacements of the grains. Redder grains indicate more displacement falling away from the initial position as they reach equilibrium owing to unconstrained sides.}
  \label{fig:extra_results}
\end{figure}

Concretely, given two adjacent blocks \(B_1\) and \(B_2\) to be stitched along the Z-axis, the last few slices of \(B_1\) and the first few slices of \(B_2\) serve as the contextual data \(\vx_0^{\mathrm{known}}\), while the gap between the two blocks constitutes the unknown region. A binary mask \(\vm\) marks the gap voxels as unknown and the context slices from both blocks as known, with the gap size configurable as a ratio of the block dimensions. A tensor of random noise \(\vx_T\) of the same spatial dimensions as the combined region (context + gap + context) is then initialized, and the inpainting inference proceeds as described in Section~\ref{sec:generation_pipeline}. This process is repeated sequentially for each pair of adjacent blocks, progressively extending the assembly. A resulting example is shown in \autoref{fig:ballast_long.png}, while \autoref{sf1:extra_results} shows a smaller stack of grains, along with their cross section.

\begin{figure}[htb]
\centering
\begin{tikzpicture}[
    block/.style={rectangle, draw, minimum width=1.6cm, minimum height=1.2cm, align=center, font=\small},
    gapstyle/.style={rectangle, draw, dashed, fill=green!20, minimum width=0.75cm, minimum height=1.2cm, align=center, font=\small},
    ctxstyle/.style={rectangle, draw, fill=yellow!45, minimum width=0.30cm, minimum height=1.2cm, align=center},
    arrow/.style={-Stealth, thick},
    lbl/.style={font=\footnotesize, text=black!70}
]

\coordinate (p1) at (0,0);
\coordinate (p2) at (5.2,0);
\coordinate (p3) at (10.45,0);

\node[block, fill=blue!15, anchor=west] (b1) at (p1) {Block\\$B_1$};
\node[block, fill=orange!15, anchor=west] (b2) at ($(p1)+(2.6,0)$) {Block\\$B_2$};
\node[lbl, above=0.2cm of $(b1.north)!0.5!(b2.north)$] {(a) Blocks with gap};
\draw[<->, dashed, thick, gray] (b1.east) -- node[above, font=\scriptsize, text=gray] {gap} (b2.west);

\draw[arrow] ($(b2.east)+(0.25,0)$) -- ($(p2)+(-0.25,0)$);

\node[block, fill=blue!15, anchor=west] (b1o) at (p2) {$B_1$};
\node[gapstyle, anchor=west] (mask) at (b1o.east) {Mask\\$\vm$};
\node[block, fill=orange!15, anchor=west] (b2o) at (mask.east) {$B_2$};
\node[ctxstyle, anchor=east] (ctxL) at (b1o.east) {};
\node[ctxstyle, anchor=west] (ctxR) at (b2o.west) {};
\node[font=\tiny, rotate=90] at (ctxL.center) {ctx};
\node[font=\tiny, rotate=90] at (ctxR.center) {ctx};
\node[lbl, above=0.2cm of mask.north] {(b) Gap masked with edge context};

\draw[arrow] ($(b2o.east)+(0.25,0)$) -- ($(p3)+(-0.25,0)$);

\node[block, fill=blue!15, anchor=west] (b1f) at (p3) {$B_1$};
\node[rectangle, draw, fill=green!40, minimum width=0.75cm, minimum height=1.2cm, align=center, font=\small, anchor=west] (filled) at (b1f.east) {Filled};
\node[block, fill=orange!15, anchor=west] (b2f) at (filled.east) {$B_2$};
\node[lbl, above=0.2cm of filled.north] {(c) After inpainting};

\end{tikzpicture}
\caption{Block stitching strategy. (a) Two independently generated blocks \(B_1\) and \(B_2\) are placed with a gap between them. (b) A binary mask \(\vm\) marks the gap region as unknown, while the edges of both blocks provide known context. (c) The inpainting model fills the gap conditioned on the context from both blocks, producing a seamless transition.}
\label{fig:stitching_schematic}
\end{figure}

\subsection{Post-processing: Segmentation}\label{sec:segmentation}
The output from the generative pipeline is a 3D binary voxel grid $V \in \{0,1\}^{N_x \times N_y \times N_z}$ representing the granular material, where $V(\vect{p})=1$ indicates a grain voxel and $V(\vect{p})=0$ indicates void space. To delineate individual grains within this dense assembly, a watershed segmentation algorithm \citep{meyer1990morphological} is employed.

The segmentation proceeds in four steps (illustrated in Figure~\ref{fig:watershed_schematic}):
\begin{enumerate}
  \item \textbf{Distance transform}: For each grain voxel $\vect{p}$, the Euclidean distance transform computes the distance to the nearest void voxel (Equation~\ref{eqn:distance_transform}):
  \begin{equation}
    D(\vect{p}) = \min_{\vect{q} : V(\vect{q})=0} \|\vect{p} - \vect{q}\|_2
    \label{eqn:distance_transform}
  \end{equation}
  Regions with high $D$ values correspond to grain centers, far from boundaries.

  \item \textbf{Marker detection}: Local maxima of $D(\vect{p})$ are identified as seed markers $\{M_1, M_2, \ldots, M_K\}$, each representing a potential grain center.

  \item \textbf{Watershed flooding}: Starting from the markers, a flooding process propagates through the distance map. Each voxel is assigned to the basin of its nearest marker, with watershed lines forming at basin boundaries where different floods meet.

  \item \textbf{Grain extraction}: Each basin defines a distinct grain with a unique label. The labeled voxels are converted to polygonal meshes via marching cubes, followed by optional erosion (radius $r_e$) and smoothing to mitigate voxelization artifacts.
\end{enumerate}

\begin{figure}[htp]
  \centering
  \includegraphics[width=0.99\textwidth]{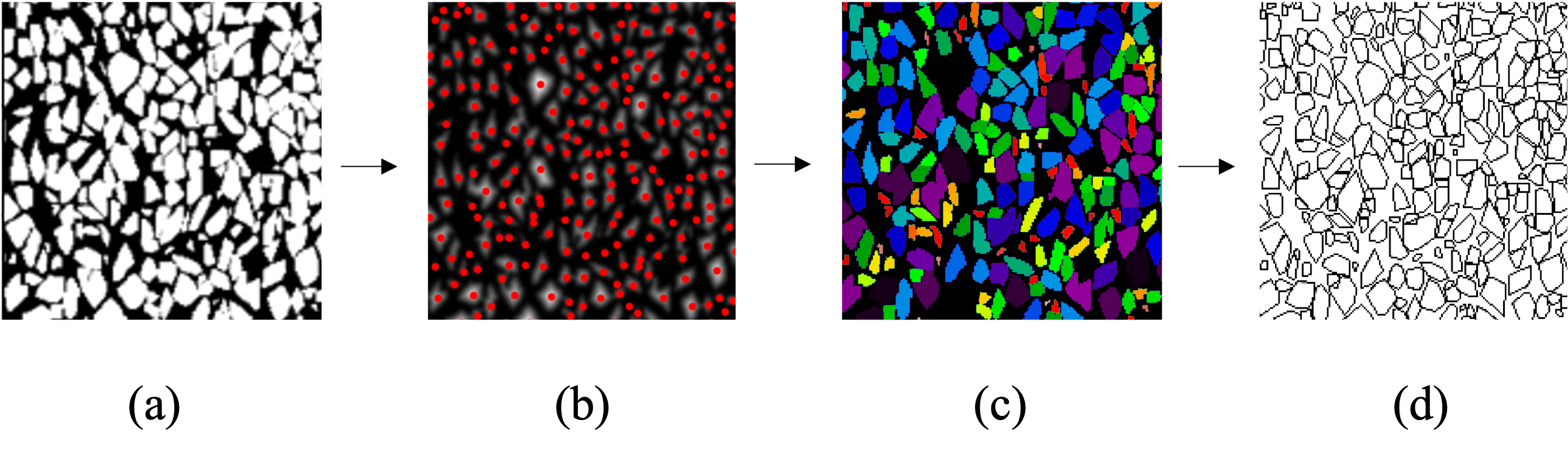}
  \caption{Watershed segmentation pipeline: (a) distance transform showing grain centers as bright regions, (b) detected local maxima serving as watershed markers, (c) watershed flooding with basin boundaries, and (d) final segmented grains delineated into polygonal meshes.}
  \label{fig:Segmentation_water_shed.png}
\end{figure}

The erosion operation shrinks grain boundaries by a parameterized radius $r_e$, computed as:
\begin{equation}
  V_{\text{eroded}}(\vect{p}) = \begin{cases} 1 & \text{if } D(\vect{p}) > r_e \\ 0 & \text{otherwise} \end{cases}
\end{equation}
This prevents inter-grain contact artifacts and enables clean convex-hull extraction for DEM compatibility. An example of this segmentation pipeline applied to a ballast sample is shown in Figure~\ref{fig:Segmentation_water_shed.png}. The output of the pipeline can be directly input into DEM simulation software such as LMGC90 \citep{fredericduboisLMGC902013} as can be seen in the image.

\begin{figure}[htb]
\centering
\begin{tikzpicture}[
    box/.style={rectangle, draw, minimum width=1.8cm, minimum height=1.2cm, align=center, font=\small},
    arrow/.style={-Stealth, thick}
]
\node[box, fill=blue!10] (binary) {Binary Grid\\$V(\vect{p}) \in \{0,1\}$};
\node[box, fill=green!10, right=0.6cm of binary] (dist) {Distance\\Transform $D$};
\node[box, fill=orange!10, right=0.6cm of dist] (markers) {Local Maxima\\(Markers)};
\node[box, fill=purple!10, right=0.6cm of markers] (watershed) {Watershed\\Flooding};
\node[box, fill=red!10, right=0.6cm of watershed] (grains) {Labeled\\Grains};

\draw[arrow] (binary) -- (dist);
\draw[arrow] (dist) -- (markers);
\draw[arrow] (markers) -- (watershed);
\draw[arrow] (watershed) -- (grains);
\end{tikzpicture}
\caption{Schematic of the watershed segmentation pipeline: binary voxel grid $\rightarrow$ distance transform $\rightarrow$ marker detection $\rightarrow$ watershed flooding $\rightarrow$ labeled grain extraction.}
\label{fig:watershed_schematic}
\end{figure}

With the methodology established, we now validate the pipeline against DEM simulations across two distinct granular materials.

\section{Validation Against DEM Samples}\label{sec:results}
For validation of the method against DEM simulation, this paper follows a set of metrics that are well-established descriptors of granular media behavior. Fabric anisotropy, coordination number (1st and 2nd order) and contact normal distributions are classical indicators of stiffness and strength distributions in granular assemblies \citep{kuhnDeformationMeasuresGranular2024}. Packing fraction and contact fabric are also known to be variables that govern the continuum behavior \citep{wenRelationVoidRatio2022}. Similarly, particle shape has importance in influencing coordination numbers and packing densities (and consequently are a proxy for stiffness) according to \citep{zhaoEffectsParticleShape2023}. Fabric anisotropy also governs variations in shear modulus according to \citep{yangEffectsFabricAnisotropy2025}.

\begin{figure}[htp]
  \centering
  \includegraphics[width=0.77\textwidth]{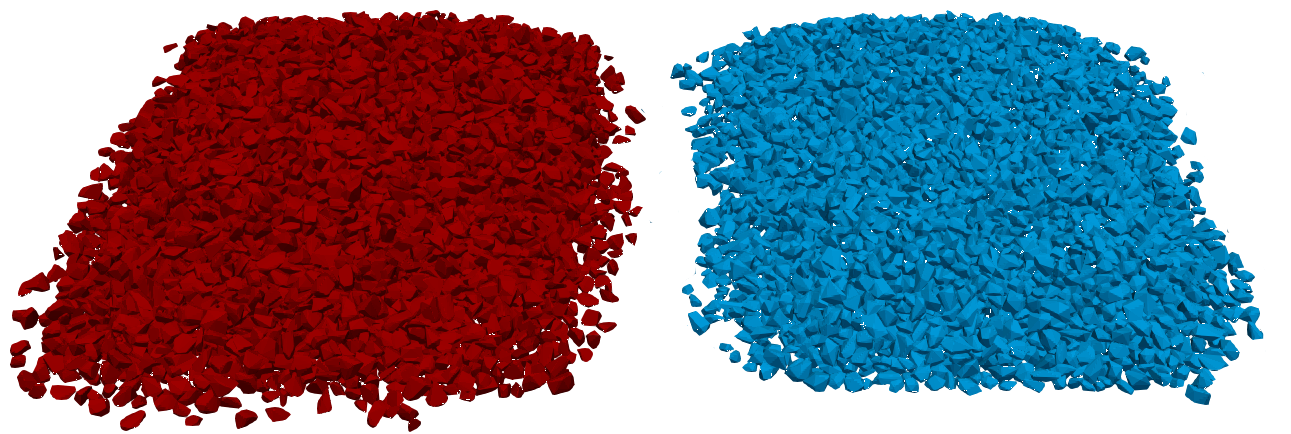}
  \caption{Left, in red: DEM generated stack of grains (7.5k particles); Right, in blue: diffusion-generated stack of grains. Both assemblies show statistically consistent coordination numbers, packing ratios, and other contact related quantities.}
  \label{fig:stack_compare.png}
\end{figure}

The pipeline was validated on two granular materials, namely railway ballast (coarse, 25--50 mm) and lunar regolith simulants (fine, sub-millimeter). Both datasets share the same preprocessing and architecture, but each material uses independently trained models (no transfer learning between ballast and regolith; see Section~\ref{sec:lunar_dataset}). Metrics were computed on assemblies with approximately 10000 particles (\autoref{fig:stack_compare.png}). We report macroscopic (packing fraction, coordination numbers, stiffness proxy), mesoscopic (fabric anisotropy, contact orientations), and microscopic (grain size, shape) statistics. Cohen's effect size (\(d = ((\mu_1 - \mu_2))/\sqrt{ ((n_1 - 1) s_1^2 + (n_2 - 1)s_2^2) / (n_1 + n_2 - 2) }\)) and $p$-values (t-test given by \(t = (\mu_1 - \mu_2)/ \sqrt{s_1^2 /n_2 + s_2^2/n_2}\), where \(\mu_i\) is the mean, \(s_i\) is the standard deviation, and \(n_i\) is the number of observations) are included to quantify differences between datasets.

\subsection{Macroscopic and granulometry statistics}\label{sec:macro_stats_section}

The fabric tensor quantifies the spatial distribution and orientation of contacts in a granular assembly. Following the classical definition in granular mechanics, the fabric tensor $\mathbf{F}$ is computed from the unit contact normal vectors as:
\begin{align}
\underset{i,j \in \{1,2,3\}}{F_{ij}} = \frac{1}{N_c} \sum_{k=1}^{N_c} n_i^{(k)} n_j^{(k)}
\end{align}
where $N_c$ is the total number of contacts, and $\vect{n}^{(k)} = (n_1^{(k)}, n_2^{(k)}, n_3^{(k)})$ is the unit normal vector at contact $k$. The anisotropy can mainly be characteristized by $\sigma_F$ and is defined as the standard deviation of the eigenvalues:
\begin{align}
  \sigma_F = \sqrt{\frac{1}{3}\sum_{i=1}^{3}(\lambda_i - \bar{\lambda})^2}
\end{align}
where $\bar{\lambda} = (\lambda_1 + \lambda_2 + \lambda_3)/3 = 1/3$. A perfectly isotropic fabric has $\sigma_F = 0$, while higher values indicate stronger directional preferences in the contact network.

\begin{table}[htb]
  \centering
  \caption{Macroscopic state and granulometry comparison between original DEM database and diffusion-generated assemblies (railway ballast example). Relative differences are computed with respect to the original DEM database.}
  \label{tab:macro_stats}
  \small
  \begin{tabular}{lccc}
    \toprule
    Metric & Original DEM & Generated & Rel. diff. [\%] \\
    \midrule
    Packing fraction ($\phi$) & 0.523 & 0.514 & -1.7 \\
    First-order coordination (mean Z1) & 5.32 & 5.29 & -0.7 \\
    Second-order coordination (mean Z2) & 17.11 & 17.66 & +3.3 \\
    Fabric tensor anisotropy ($\sigma_F$) & 0.195 & 0.171 & -12.3 \\
    Mechanical stiffness proxy ($\propto Z^2$) & 1.00 & 0.99 & -1.4 \\
    \bottomrule
  \end{tabular}
\end{table}

Table~\ref{tab:macro_stats} summarizes these macroscopic comparisons. The generated packing fraction remains close to the DEM baseline ($-1.7\%$ relative difference). The mean first-order coordination number is statistically indistinguishable from the DEM reference (Z1: $-0.7\%$, Cohen's $d=0.018$, $p=0.27$), while the second-order coordination shows a small increase (Z2: $+3.3\%$, $d=-0.079$, $p < 10^{-5}$). Fabric tensor anisotropy is lower in generated samples by $12.3\%$, while the mechanical stiffness proxy decreases by only $1.4\%$. Grain-size distributions show close agreement (Kolmogorov-Smirnov test: $\text{KS}=0.043$, $p < 10^{-5}$). Detailed distribution and shape comparisons are presented in \autoref{fig:stats_dashboards}. A detailed explanation of the metrics is presented in \autoref{sec:metric_definitions}.

\subsection{Particle shape and fabric metrics}\label{sec:particle_shape_fabric}

\begin{figure*}[htp]
  \centering
  \begin{subfigure}[t]{0.32\textwidth}
    \includegraphics[width=\textwidth]{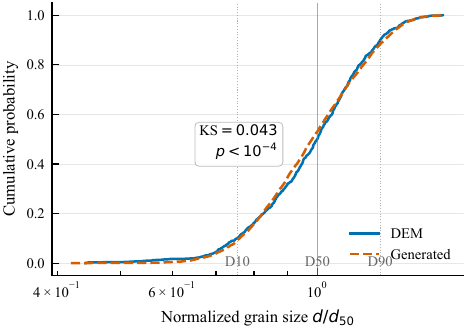}
    \caption{Grain size CDF}\label{fig:grainsize_cdf}
  \end{subfigure}\hfill
  \begin{subfigure}[t]{0.32\textwidth}
    \includegraphics[width=\textwidth]{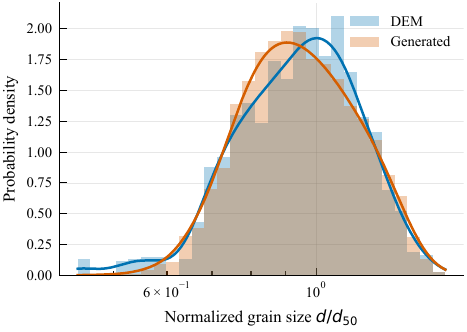}
    \caption{Grain size PDF}\label{fig:grainsize_pdf}
  \end{subfigure}\hfill
  \begin{subfigure}[t]{0.32\textwidth}
    \includegraphics[width=\textwidth]{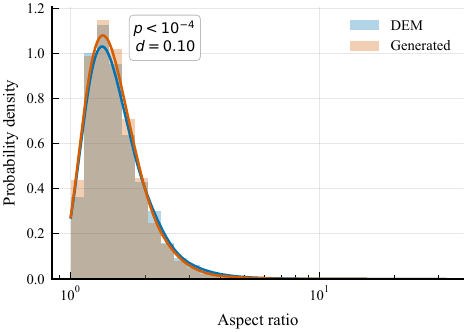}
    \caption{Aspect ratio}\label{fig:aspect_ratio}
  \end{subfigure}

  \medskip

  \begin{subfigure}[t]{0.32\textwidth}
    \includegraphics[width=\textwidth]{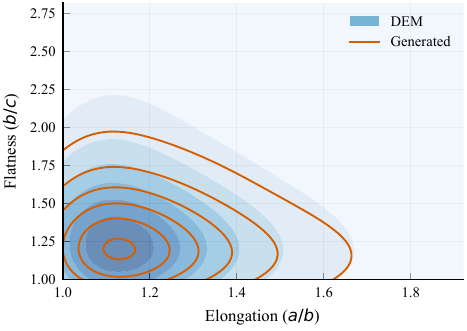}
    \caption{Elongation vs.\ flatness (KDE)}\label{fig:elong_flat}
  \end{subfigure}\hfill
  \begin{subfigure}[t]{0.32\textwidth}
    \includegraphics[width=\textwidth]{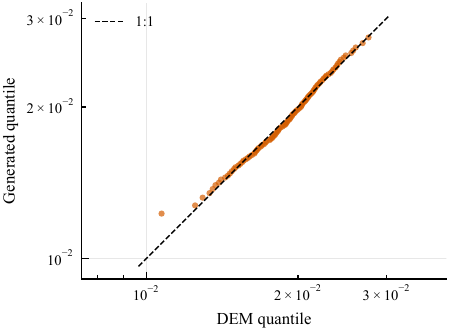}
    \caption{Grain size Q--Q}\label{fig:grainsize_qq}
  \end{subfigure}\hfill
  \begin{subfigure}[t]{0.32\textwidth}
    \includegraphics[width=\textwidth]{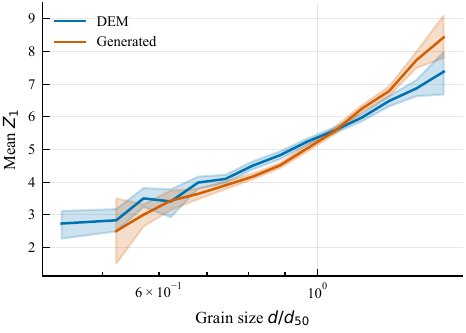}
    \caption{$Z_1$ vs.\ grain size}\label{fig:coord_size}
  \end{subfigure}
  \caption{Granulometry and shape validation: (a)~cumulative and (b)~probability density of normalized grain sizes, (c)~aspect ratio distribution, (d)~elongation vs.\ flatness joint density (KDE contours), (e)~quantile--quantile plot, and (f)~binned mean coordination vs.\ grain size with 95\% confidence interval bands. DEM reference in blue, diffusion generated in orange.}
  \label{fig:stats_dashboards}
\end{figure*}

\begin{figure*}[htp]
  \centering
  \begin{subfigure}[t]{0.32\textwidth}
    \includegraphics[width=\textwidth]{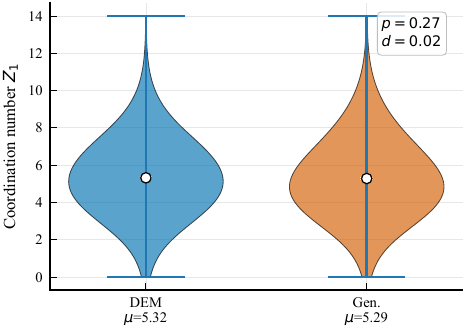}
    \caption{$Z_1$ violin}\label{fig:z1_violin}
  \end{subfigure}\hfill
  \begin{subfigure}[t]{0.32\textwidth}
    \includegraphics[width=\textwidth]{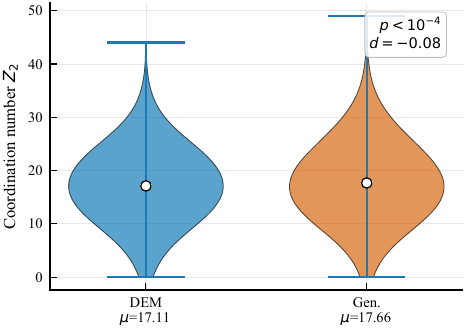}
    \caption{$Z_2$ violin}\label{fig:z2_violin}
  \end{subfigure}\hfill
  \begin{subfigure}[t]{0.32\textwidth}
    \includegraphics[width=\textwidth]{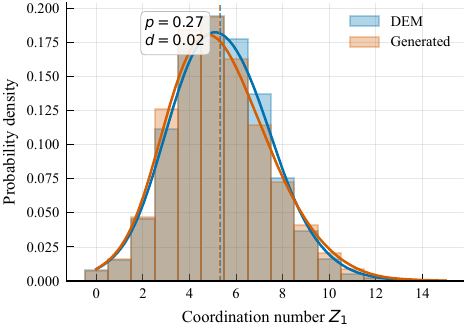}
    \caption{$Z_1$ distribution}\label{fig:z1_dist}
  \end{subfigure}

  \medskip

  \begin{subfigure}[t]{0.32\textwidth}
    \includegraphics[width=\textwidth]{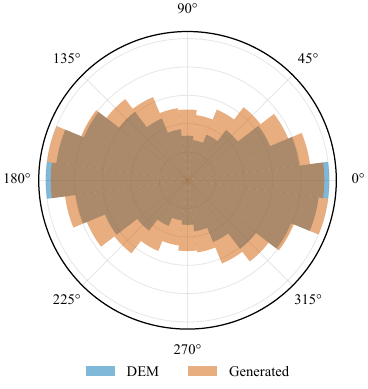}
    \caption{Rose diagram (XY)}\label{fig:rose_xy}
  \end{subfigure}\hfill
  \begin{subfigure}[t]{0.32\textwidth}
    \includegraphics[width=\textwidth]{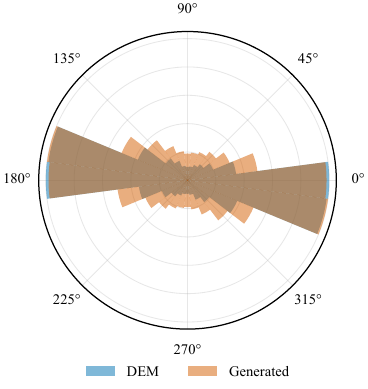}
    \caption{Rose diagram (XZ)}\label{fig:rose_xz}
  \end{subfigure}\hfill
  \begin{subfigure}[t]{0.32\textwidth}
    \includegraphics[width=\textwidth]{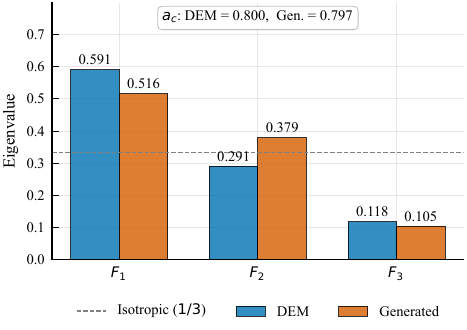}
    \caption{Fabric tensor eigenvalues}\label{fig:fabric_anis}
  \end{subfigure}
  \caption{Mechanical and fabric validation: (a,\,b)~first and second-order coordination number violins with 95\% confidence interval bands, (c)~$Z_1$ histogram, (d,\,e)~contact-normal rose diagrams in the XY and XZ planes, and (f)~fabric tensor eigenvalue decomposition ($F_1 \geq F_2 \geq F_3$) with isotropic reference. DEM reference in blue, diffusion-generated in orange.}
  \label{fig:mech_dashboards}
\end{figure*}

For railway ballast, the rose diagrams, elongation--flatness maps, and aspect-ratio distributions remain close to DEM references (\autoref{fig:mech_dashboards}, \autoref{fig:stats_dashboards}). The fabric tensor anisotropy coefficient is slightly lower in generated samples (0.171 vs 0.195), indicating a marginally more isotropic fabric. Contact-normal distributions and shape-proxy scatter plots remain consistent with DEM trends.

The generated assemblies exhibit negligible deviations in coordination numbers, with the first-order coordination statistically indistinguishable from the DEM reference (Cohen's $d = 0.018$, $p = 0.27$). The effect sizes remain small (Cohen's $|d| < 0.08$), suggesting that the deviations are of little mechanical importance. The fabric tensor anistropy differs by -12.3\%, but considering their absolute values are very small (almost zero, showing that both samples are isotropic), the deviation carries no physical meaning and is a numerical artifact.

\subsection{Inpainting seam quality metrics}

\begin{figure*}[htp]
  \centering
  \begin{subfigure}[t]{0.32\textwidth}
    \includegraphics[width=\textwidth]{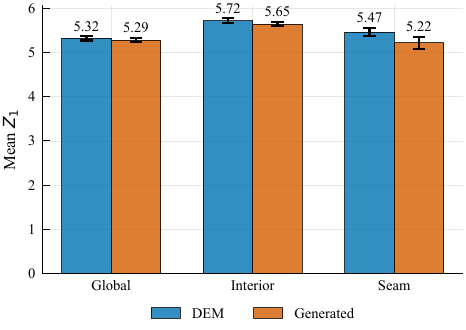}
    \caption{Regional $Z_1$ comparison}\label{fig:region_z1}
  \end{subfigure}\hfill
  \begin{subfigure}[t]{0.32\textwidth}
    \includegraphics[width=\textwidth]{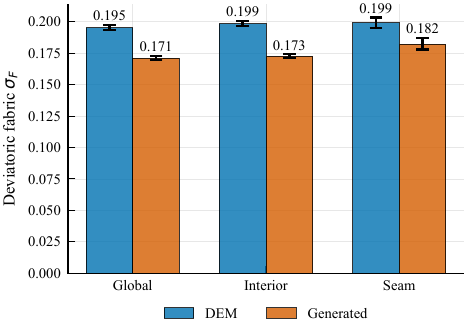}
    \caption{Regional fabric $\sigma_F$}\label{fig:region_fabric}
  \end{subfigure}\hfill
  \begin{subfigure}[t]{0.32\textwidth}
    \includegraphics[width=\textwidth]{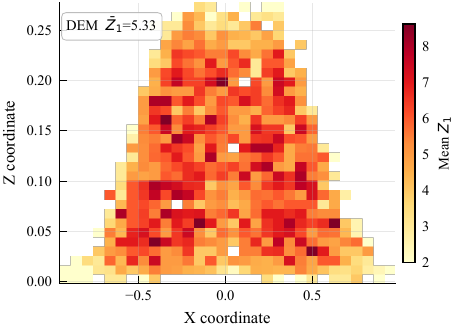}
    \caption{DEM mid-plane $Z_1$ map}\label{fig:slice_dem}
  \end{subfigure}

  \medskip

  \begin{subfigure}[t]{0.32\textwidth}
    \includegraphics[width=\textwidth]{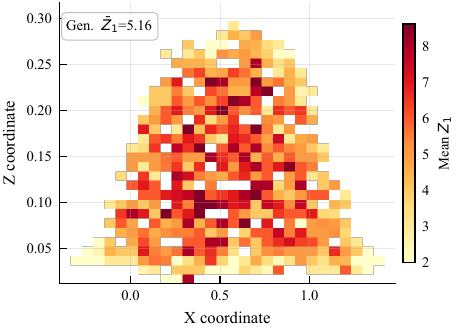}
    \caption{Generated mid-plane $Z_1$ map}\label{fig:slice_gen}
  \end{subfigure}\hfill
  \begin{subfigure}[t]{0.32\textwidth}
    \includegraphics[width=\textwidth]{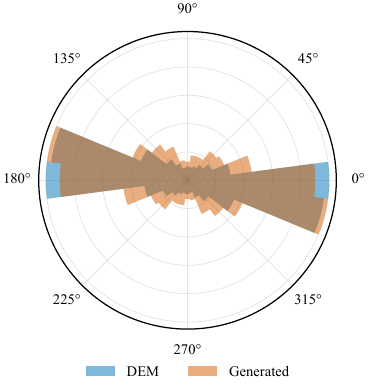}
    \caption{Seam rose diagram (XZ)}\label{fig:seam_rose}
  \end{subfigure}\hfill
  \begin{subfigure}[t]{0.32\textwidth}
    \includegraphics[width=\textwidth]{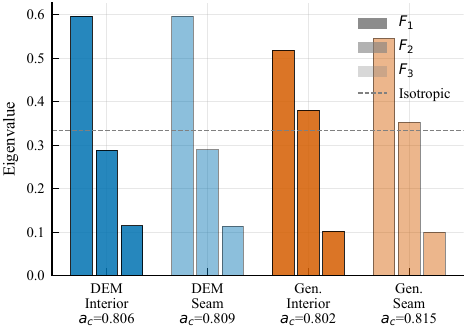}
    \caption{Seam vs.\ interior eigenvalues}\label{fig:seam_eig}
  \end{subfigure}
  \caption{Seam quality assessment: (a)~mean coordination number $Z_1$ by region (global, interior, seam) with 95\% confidence interval bands, (b)~deviatoric fabric magnitude $\sigma_F$ by region, (c,\,d)~mid-plane $Z_1$ heatmaps for DEM and generated assemblies, (e)~contact-normal rose diagram (XZ) for seam contacts only, and (f)~fabric tensor eigenvalue decomposition for interior vs.\ seam regions. DEM reference in blue, diffusion generated in orange.}
  \label{fig:seam_quality}
\end{figure*}

To assess the quality of the inpainting process used to stitch together independently generated blocks, seam and interior regions were analyzed separately. We compare first-order coordination and fabric anisotropy, as summarized in Table~\ref{tab:seam_metrics} and \autoref{fig:seam_quality}.

\begin{table}[htb]
  \centering
  \caption{Comparison of metrics in seam vs interior regions (railway ballast example). For seam regions, equivalent areas from the DEM are used to compute the metrics, hence slightly varying values from the assembly averages.}
  \label{tab:seam_metrics}
  \begin{tabular}{lccc}
    \toprule
    Region & Metric & DEM & Generated \\
    \midrule
    Assembly (overall) & Z1 & 5.32 & 5.29 \\
    Assembly (overall) & $\sigma_F$ & 0.195 & 0.171 \\
    \midrule
    Seam & Z1 & 5.47 & 5.22 \\
    Seam & $\sigma_F$ & 0.199 & 0.182 \\
    \midrule
    Interior & Z1 & 5.72 & 5.65 \\
    Interior & $\sigma_F$ & 0.199 & 0.173 \\
    \bottomrule
  \end{tabular}
\end{table}

The seam regions show coordination numbers very close to interior values in both assemblies (DEM seam: Z1 = 5.47 vs interior: 5.72; generated seam: Z1 = 5.22 vs interior: 5.65). This indicates that inpainting produces physically consistent transitions without introducing contact-network discontinuities. The fabric anisotropy $\sigma_F$ is also comparable between seam and interior for both DEM (0.199 vs 0.199) and generated samples (0.182 vs 0.173). For generated samples, the small residual differences are consistent with the seam-mask geometry used during inpainting.

\begin{figure}[htp]
  \centering
  \includegraphics[width=0.99\textwidth]{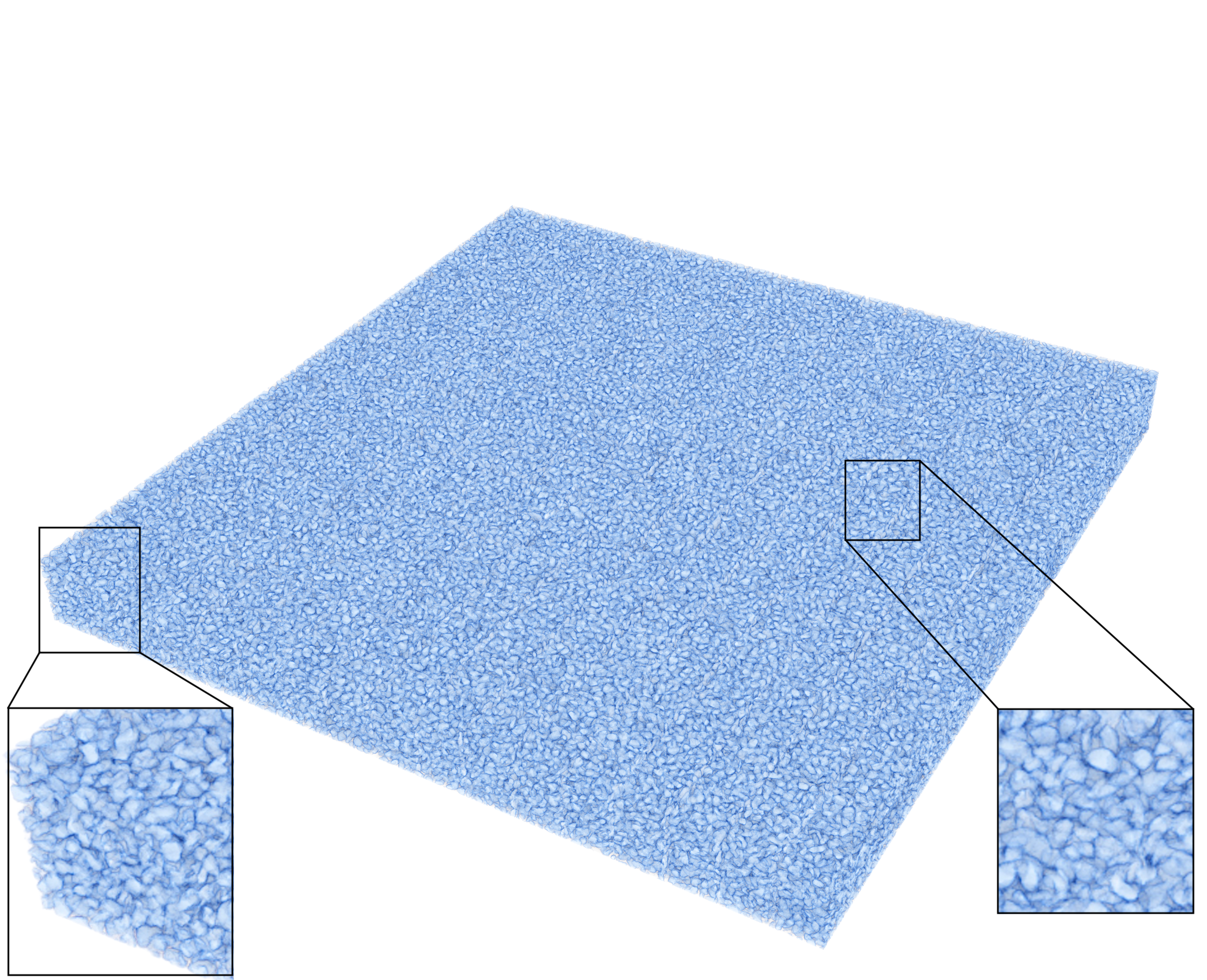}
  \caption{Large-scale lunar regolith surface generated by the diffusion pipeline. The sample is a $34 \times 34 \times 2$ arrangement of $32 \times 64 \times 64$ voxel blocks (approximately 5--7 cm per dimension) representing a compacted lunar regolith simulant surface. Each block took less than 3 seconds to generate; with 16 blocks in parallel, total generation time was approximately 480 seconds including inpainting.}
  \label{fig:lunar_regolith_example.png}
\end{figure}

For lunar regolith, inpainting also produces seamless transitions between blocks, enabling continuous terrain surfaces (\autoref{fig:lunar_regolith_example.png}) without visible boundary artifacts.

\subsection{Computational Performance}
The inpainting model expands the UNet input from one to three channels (noisy latent, binary mask, known region), which increases compute by roughly $\sim2.5\times$ and peak memory by $\sim3.5\times$ relative to unconditional generation. A variant that denoises only masked voxels reduces this overhead to $\sim1.3\times$ without noticeable artifacts. In practice, unconditional synthesis runs with a batch size of $64$ for $32\!\times\!64\!\times\!64$ grids, while inpainting schedules up to $32$ seams in parallel.

\begin{figure*}[htp]
  \centering
  \includegraphics[width=0.85\textwidth]{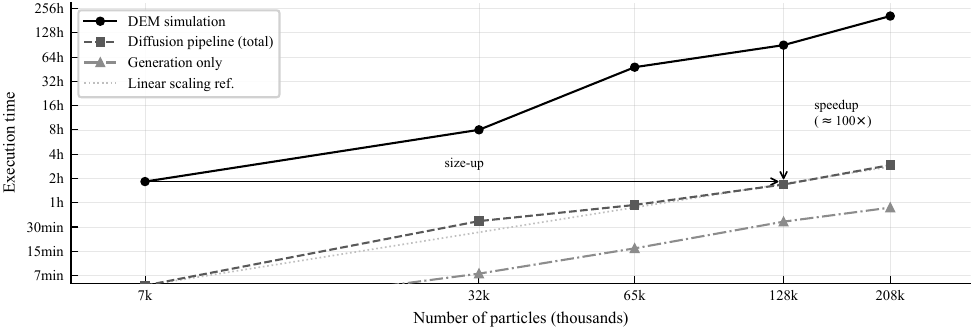}
  \caption{Scalability comparison between DEM simulation and the diffusion pipeline. The plot shows execution time (log scale) versus number of particles. Circles (solid): full DEM simulation time on CPU; squares (dashed): total diffusion pipeline time (GPU generation + CPU segmentation + DEM post-processing); triangles (dash-dot): diffusion generation time only on GPU. The vertical ``speedup'' arrow illustrates the $\sim$100$\times$ time reduction for the same problem size, while the horizontal ``size-up'' arrow shows that larger problems can be solved in the same time budget. Experiments performed on Jean Zay \autoref{ack} cluster (1$\times$H100 GPU, 4 CPU cores for segmentation, 4 CPU cores for DEM-Post Processing and full DEM simulations).}
  \label{fig:DEMvsDDPMspeedup.png}
\end{figure*}

For a long sample of size of approximately 200{,}000 particles, diffusion generation took about $50$ minutes on one NVIDIA H100 GPU, and segmentation took about another $65$ minutes on $4$ Intel Xeon CPU cores. On the same CPU configuration, DEM initialization for comparable sizes (above 128k particles) exceeded 150 hours and increased rapidly with particle count. A detailed breakdown of speedup curves in \autoref{fig:DEMvsDDPMspeedup.png} therefore show three quantities separately (full DEM runtime (CPU), diffusion generation only (GPU), and diffusion generation + segmentation + DEM post processing).

In short, the speedup for a typical example is more than 100x, and becomes more pronounced with increasing particle count. This becomes significant when generating large datasets. For perspective, a dataset of 300 examples of 30\,m ballast tracks was generated within 24 hours; meanwhile, on the same platform, complete DEM simulation could not produce a single 30\,m example in that time window. For reproducibility, hardware usage was fixed as follows: diffusion generation on 4$\times$NVIDIA H100 GPUs (with 4-way inpainting parallelism), and DEM simulation/segmentation on 16$\times$Intel Xeon CPU cores.

For lunar regolith, similar speedups were observed. The diffusion pipeline generates equivalent volumes in seconds, representing speedups exceeding 100$\times$. \autoref{fig:lunar_regolith_example.png} shows a large-scale lunar surface (a $34 \times 34 \times 2$ arrangement of voxel blocks, approximately 5--7 cm per dimension) generated in under 8 minutes including inpainting time. This ``Infinite Lunar Regolith'' capability without requiring particle by particle deposition enable simulations of "lunar" scenarios, most notably Rover simulations on long terrains \citep{hurrellLunarRoverDiscrete2023}.

These results demonstrate that the diffusion pipeline successfully generates physically plausible granular assemblies across materials with fundamentally different characteristics. The implications and limitations of these findings are now discussed.

\section{Current Limitations and Future Directions}\label{sec:discussion}

While the pipeline demonstrates strong agreement with DEM baselines, a few limitations are acknowledged:

\begin{enumerate}
  \item \textbf{Dependency on training data}: The quality and diversity of generated samples is fundamentally limited by the DEM training data. If the source simulations do not cover certain physical states or material properties, the generative model does not reliably generalize to these unseen states without additional retraining or transfer learning.

  \item \textbf{Voxel resolution constraints}: The current implementation uses $32 \times 64 \times 64$ voxel blocks, which imposes a tradeoff between memory consumption and geometric fidelity. Higher resolutions would better capture fine grain boundary details but require significantly more GPU memory.

  \item \textbf{Isotropy assumption}: The inpainting process assumes approximate isotropy in the voxel grids. While gravity-induced anisotropy in the vertical direction is small for the block sizes considered, this assumption may not hold for materials with strong directional preferences. There is a preliminary support for producing anisotropic blocks unconditionally, showing an ability of the same architecture to produce such anisotropy, but further statistical validation is left for a future work.

  \item \textbf{Segmentation accuracy}: The watershed segmentation can struggle with touching or overlapping grains, potentially affecting coordination number statistics. The erosion parameter provides some control but requires careful tuning.
\end{enumerate}

Implementing a faster segmentation method (the process that is the slowest step in the pipeline), is not trivial and would require significant extension to the work. Nevertheless, the current segmentation pipeline is still only a fraction of the total computation time a similar DEM simulation would take. Future work could explore the use of learned segmentation methods, potentially reducing inference steps and improving grain boundary accuracy.

Several avenues for future work emerge from this study. Conditional generation could target specific physical states (e.g., desired packing fraction or coordination number) rather than sampling from the learned distribution. Although, the model itself can generate non-convex particles, the segmentation pipleine limits the particles to locally convex shapes only. Extension to non-convex grain geometries would broaden applicability to materials like crushed rock or biological particles. Transfer learning between material types (e.g., initializing regolith models from ballast weights) could reduce training data requirements and accelerate adaptation to new granular materials. Finally, integration with physics-informed loss functions during training could further improve mechanical consistency with DEM simulations.

\section{Conclusion}\label{sec:conclusion}

This paper presents a 3D diffusion-based pipeline for generating large-scale granular assemblies that are statistically and mechanically consistent with DEM simulations. The scale of the simulations produced are traditionally impossible to achieve with lengthy DEM simulations. The core contribution is a two-stage neural network architecture and generation strategy. It begins with an unconditional 3D diffusion model that generates independent voxel blocks representing packed granular media, and follows after with a 3D inpainting model that seamlessly joins these blocks by conditioning on masked boundary regions. The inpainting model combines channel-based mask conditioning with RePaint-style guidance, enabling the generation of arbitrarily large, continuous granular volumes without any stitching artifacts.

The methodology was validated on two distinct applications (railway ballast and lunar regolith simulants). Quantitative comparisons demonstrate close agreement with the original DEM database across multiple physical metrics including coordination numbers, fabric tensor anisotropy, grain size distributions, and packing fractions. The pipeline achieves speedups exceeding 100x compared to equivalent DEM simulations, with the advantage becoming more pronounced as assembly size increases. For perspective, a 30-meter railway track that would require days of DEM computation can be generated in under an hour. More importantly, the pipeline enables simulations of upwards of 1 million particles in the same time a DEM simulation would only complete a sub 50,000 particles simulation.

Beyond computational efficiency, the pipeline offers several practical advantages. It learns a generative distribution over granular configurations, scales linearly with output size, and can be quickly retrained on new granular media samples. The pipeline represents a practical tool for industrial applications requiring rapid generation of realistic granular assemblies. In the future, learned segmentation methods could replace watershed-based post-processing, potentially reducing inference steps and improving grain boundary accuracy.

\begin{ack}
\label{ack}

This work is done under a PhD project funded by SNCF, France (CIFRE: N° 2023/1116). This work was granted access to the HPC resources of IDRIS under the allocation 2025-AD011016527 made by GENCI. This work also granted access to the HPC resources of the Copernicus cluster at Université Aix Marseille under the allocation ``A441''.
\end{ack}

\section*{Conflict of Interest}
The authors declare no conflicts of interest.

\section*{Author Contributions}
\textbf{Muhammad M. Hassan}: Methodology, conceptualization, writing the manuscript.
\textbf{Régis Cottereau}: Supervision, conceptualization, reviewing and editing.
\textbf{Filippo Gatti}: Supervision, conceptualization, reviewing and editing.
\textbf{Patryk Dec}: Supervision, conceptualization, reviewing and editing.

\section*{Code and Data Availability}
The code for the diffusion pipeline, including training scripts, inference routines, and segmentation tools, is available at \url{https://github.com/FammasMaz/fast-gran-gen}. The NU-LHT-4M grain shape library used for lunar regolith is publicly available from the NIST/NASA database \citep{kafkaThreeDimensionalShape2024}. Trained model weights for railway ballast and lunar regolith will be made available following publication.

\newpage

\appendix

\section{Definition of Macroscopic Metrics}\label{sec:metric_definitions}

The macroscopic metrics used in Table~\ref{tab:macro_stats} to compare original DEM and generated assemblies are defined below.

\paragraph{Packing fraction ($\phi$)}
The packing fraction (or solid volume fraction) is the ratio of the total volume occupied by solid grains $V_s$ to the total volume of the assembly domain $V$:
\begin{align}
  \phi = \frac{V_s}{V} = \frac{\sum_{i=1}^{N_p} V_i}{V}
\end{align}
where $N_p$ is the number of particles and $V_i$ is the volume of particle $i$. In the voxelized representation, $V_s$ is computed by counting all non-void voxels.

\paragraph{First-order coordination number ($Z_1$)}
The first-order (or direct) coordination number of a particle is the number of distinct particles with which it shares at least one contact. The mean first-order coordination number is:
\begin{align}
  Z_1 = \frac{1}{N_p}\sum_{i=1}^{N_p} z_1^{(i)}
\end{align}
where $z_1^{(i)}$ is the number of direct contact neighbours of particle $i$.

\paragraph{Second-order coordination number ($Z_2$)}
The second-order coordination number counts the number of distinct particles reachable within two contact hops from a given particle (excluding the particle itself and its direct neighbours). Its mean is:
\begin{align}
  Z_2 = \frac{1}{N_p}\sum_{i=1}^{N_p} z_2^{(i)}
\end{align}
where $z_2^{(i)} = \left|\bigcup_{j \in \mathcal{N}_1(i)} \mathcal{N}_1(j) \;\setminus\; (\{i\} \cup \mathcal{N}_1(i))\right|$ and $\mathcal{N}_1(i)$ denotes the set of direct contact neighbours of particle $i$. This metric characterizes the medium-range connectivity of the granular network.

\paragraph{Fabric tensor anisotropy ($\sigma_F$)}
The fabric tensor $\mathbf{F}$ characterizes the directional distribution of contacts (see \autoref{sec:macro_stats_section} for its full definition). Its anisotropy is quantified by the standard deviation of its eigenvalues $\lambda_1, \lambda_2, \lambda_3$:
\begin{align}
  \sigma_F = \sqrt{\frac{1}{3}\sum_{i=1}^{3}(\lambda_i - \bar{\lambda})^2}, \qquad \bar{\lambda} = \frac{1}{3}
\end{align}
A value of $\sigma_F = 0$ indicates a perfectly isotropic contact network, while larger values indicate preferential contact orientations.

\paragraph{Mechanical stiffness proxy ($\propto Z^2$)}
Under the effective medium theory for granular assemblies, the bulk modulus $K$ of a random packing of elastic spheres scales as $K \propto \phi\, Z^2$ at a given confining pressure. The stiffness proxy reported in Table~\ref{tab:macro_stats} is the normalized ratio:
\begin{align}
  S = \frac{\phi_{\mathrm{gen}}\, Z_{1,\mathrm{gen}}^2}{\phi_{\mathrm{orig}}\, Z_{1,\mathrm{orig}}^2}
\end{align}
so that $S = 1$ when the generated assembly exactly matches the original DEM database. This provides a first-order estimate of how the macroscopic elastic stiffness of the generated packing compares to that of the reference.

\section{Ablation Studies}\label{sec:ablation}
Several models across multiple techniques have been tested to generate the 3D voxels grids. Similarly multiple blending techniques were explored, which led to ultimately settling on the 3D diffusion with the inpainting model. The following subsections explore some of these tests.

\subsection{Latent diffusion models}
An immediate first choice when dealing with 3D dataset is to usually compress the dataset into a smaller latent space that accurately captures the topological characteristics of the samples in the dataset. For the ballast application explored in this example, the most important characteristics are the grain boundaries and the contact between them. As a first, a Variational Autoencoder was tested, which learnt 2D latents from the 3D dataset, convoluting multiple learnt planar views into a final 2D output. Other techniques like self-attention \citep{vaswaniAttentionAllYou2023}, and CBAM was also tested, along with a VQVAE \citep{oordNeuralDiscreteRepresentation2018}, Fourier domain filtering \citep{yuDMFFTImprovingGeneration2025}, and standard Autoencoders \citep{bankAutoencoders2021}, however even the best results obtained were produced grid artifacts. This can be seen in \autoref{fig:sdf_values.png}

\subsection{Representation of grain topology as level-sets}\label{sec:level_sets}
As an alternative to the direct binary voxel representation described in \autoref{sec:dataset_slicing}, the use of Signed Distance Fields (SDFs) or level sets was explored for representing grain topology. The motivation was to explore whether a continuous representation could enable the diffusion models to better capture the smooth surfaces of granular particles and potentially improve the fidelity of generated grain boundaries, especially when latent models were being used.

In this approach, each binary voxel grid from the dataset was converted into an SDF representation. This was achieved by computing the Euclidean distance transform for voxels inside and outside the grains, with the sign indicating interior (negative) versus exterior (positive) regions relative to a grain's surface. The resulting SDF values were then normalized, typically to a range of \(\left[ -1, 1 \right]\). The diffusion models were then be trained to directly generate these SDF volumes.

The empirical results, however, indicated that for the specific characteristics of the ballasted granular media under study, training diffusion models directly on these SDF representations did not yield a observable improvement in the quality or realism of the generated samples compared to models trained on the simpler binary voxel grids. They also needed on average 5x denser UNet backbones to generate respectable but blurry results. Albeit this largely eradicated ``grid artificats'' that were seen with the latent models, the quality with vanilla 3D models remained superior. An example of a 3D voxel generated using SDF values is shown in \autoref{fig:grid_art.jpg}.

\subsection{Alternative blending techniques}\label{sec:alt_blending}
Prior to adopting the more sophisticated inpainting-based approaches for ensuring coherency between generated 3D voxel grids, a simpler blending technique was explored. This computationally less intensive methodology was as follows:
\begin{enumerate}
    \item A series of \(N\) individual 3D voxel blocks were generated independently using the unconditional diffusion model, as described in \autoref{sec:unconditional_generation}.
    \item These blocks were then assembled sequentially. For any two adjacent blocks, if an overlap of \(k\) slices was defined, the values in this overlapping region were combined by a direct averaging method. Specifically, for each voxel position \((x,y,z)\) within the \(k\) overlapping slices, the new value \(V_{new}(x,y,z)\) was computed as \((V_{block_i}(x,y,z) + V_{block_{i+1}}(x,y,z)) / 2\).
    \item The final larger volume was constructed by joining the non-overlapping part of the first block, the averaged overlap region, and the non-overlapping part of the second block, this process being repeated for all subsequent blocks.
    \item Iterations are done by inputting the blended part as the latent for denoising and the overlapping part is kept and blended in again.
\end{enumerate}

\begin{figure}[htb]
  \centering
  \begin{subfigure}[htb]{0.33\textwidth}
    \includegraphics[width=\textwidth]{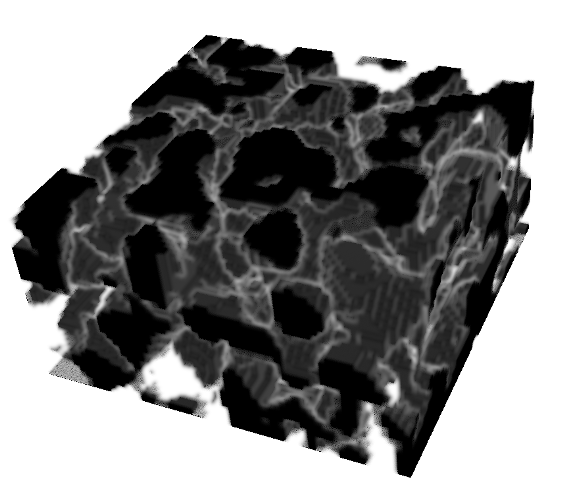}
    \caption{Blurry edges when using latent models as input}
    \label{fig:sdf_values.png}
  \end{subfigure}
  \begin{subfigure}[htb]{0.32\textwidth}
    \includegraphics[width=\textwidth]{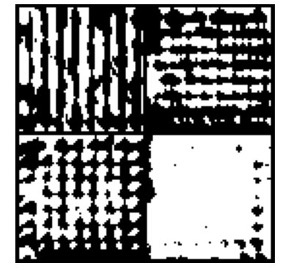}
    \caption{Grid artifacts when using SDF values in inference on 4 random 2D slices}
    \label{fig:grid_art.jpg}
  \end{subfigure}
  \begin{subfigure}[htb]{0.30\textwidth}
    \includegraphics[width=\textwidth]{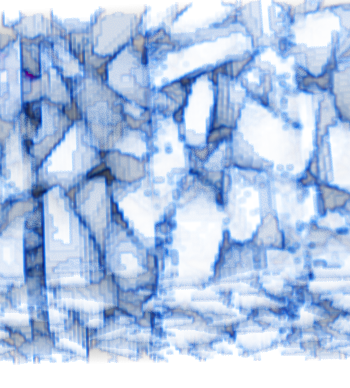}
    \caption{Noisy joints when using averaging blending}
    \label{fig:noisy_joints.jpg}
  \end{subfigure}
  \caption{Comparison of the quality of the generated samples when using latent models, SDF values as input, and averaging blending}
\end{figure}

While this averaging approach is computationally efficient and straightforward to implement, it proved to be suboptimal for generating high-quality, seamless granular assemblies. The primary drawback was the introduction of noticeable artifacts and noise in the blended regions. This can be seen in \autoref{fig:noisy_joints.jpg}, leading us to inpainting model as a better choice.

\bibliography{references}

\end{document}